\documentclass{IEEEtaes}

\usepackage{color,array,amsthm}
\usepackage{graphicx}
\usepackage{amsmath,amsfonts}
\usepackage[caption=false,font=normalsize,labelfont=sf,textfont=rm]{subfig}
\usepackage{tabularx}
\usepackage{booktabs}
\usepackage{multirow}
\usepackage{balance}

\jvol{XX}
\jnum{XX}
\jmonth{XXXXX}
\paper{1234567}
\pubyear{2022}
\doiinfo{TAES.2022.Doi Number}

\begin{document}

\title{SCT-MOT: Enhancing Air-to-Air Multiple UAVs Tracking with Swarm-Coupled Motion and Trajectory Guidance}

\author{ZHAOCHEN CHU}
%\member{Fellow, IEEE}
\affil{Beijing Institute of Technology, Beijing, China} 

\author{TAO SONG}
%\affil{Beijing Institute of Technology, Beijing, China} 

\author{REN JIN*}
%\member{Member, IEEE}
%\affil{Beijing Institute of Technology, Beijing, China}

\author{SHAOMING HE}

\author{DEFU LIN}
%\member{Member, IEEE}
\affil{Beijing Institute of Technology, Beijing, China}
\author{SIQING CHENG}
%\author{SHAOMING HE}
%\member{Member, IEEE}
%\affil{Beijing Institute of Technology, Beijing, China}

%\author{YING FU}
%\member{Member, IEEE}
\affil{ Xi'an Jiaotong-Liverpool University, Suzhou, China}

%% \author{FOURTH D. AUTHOR}
%% \affil{University of Colorado, Colorado, USA}

\receiveddate{
This study was co-supported by the National Key Research and Development Program of China (No. 2021YFF0601304) and the National Natural Science Foundation of China (No. 62206020). \\
This work has been submitted to the IEEE for possible publication. Copyright may be transferred without notice, after which this version may no longer be accessible.
}
%% \accepteddate{XXXXX XX XXXX}
%% \publisheddate{XXXXX XX XXXX}

\corresp{
%The name of the corresponding author appears after the financial information, e.g.
	 {\itshape (Corresponding author: Ren Jin)}}

\authoraddress{
Zhaochen Chu, Tao Song, Ren Jin, ShaoMing He, Defu Lin are with the China-UAE Belt and Road Joint Laboratory on Intelligent Unmanned Systems of School of Aerospace Engineering, Beijing Institute of Technology, Beijing, 100081, China, (e-mail:  \href{2315228186@qq.com}{2315228186@qq.com};  \href{6120160130@bit.edu.cn}{6120160130@bit.edu.cn};  \href{renjin@bit.edu.cn}{renjin@bit.edu.cn};  
\href{shaoming.he@bit.edu.cn}{shaoming.he@bit.edu.cn}; \href{lindf@bit.edu.cn}{lindf@bit.edu.cn}).	
Siqing Cheng is with School of Information and Computing Science, Xi'an Jiaotong-Liverpool University, Suzhou, 215123, China (e-mail:  \href{Siqing.Cheng24@student.xjtlu.edu.cn}{Siqing.Cheng24@student.xjtlu.edu.cn}).	
}

%\editor{Mentions of supplemental materials and animal/human rights statements can be included here.}
%\supplementary{Color versions of one or more of the figures in this article are available online at \href{http://ieeexplore.ieee.org}{http://ieeexplore.ieee.org}.}

\markboth{ZHAOCHEN CHU ET AL.}{SCT-MOT}
\maketitle

\begin{abstract}
Air-to-air tracking of multiple UAVs in swarm scenarios presents significant challenges due to the complex nonlinear group movements and weak per-frame visual cues for small UAV objects, which often result in detection failures, trajectory fragmentation, and identity switches. Although existing methods have attempted to improve performance by incorporating trajectory prediction, they model each object independently, neglecting the swarm-level motion dependencies. Moreover, their limited integration between motion prediction and appearance representation weakens the spatio-temporal consistency required for tracking in visually ambiguous and cluttered environments, making it difficult to maintain coherent trajectories and reliable associations.
To address these challenges, we propose SCT-MOT, a multiple UAVs tracking framework that integrates Swarm-Coupled motion modeling and Trajectory-guided feature fusion. First, we develop a Swarm Motion-Aware Trajectory Prediction (SMTP) module that jointly models historical trajectories and posture-aware appearance features from a swarm-level perspective, enabling more accurate forecasting of the nonlinear, coupled group trajectories. Second, we design a Trajectory-Guided Spatio-Temporal Feature Fusion (TG-STFF) module that aligns predicted positions with historical visual cues and deeply integrates them with current frame features, enhancing temporal consistency and spatial discriminability for weak objects. Extensive experiments on three public air-to-air swarm UAV tracking datasets, including AIRMOT, MOT-FLY, and UAVSwarm, demonstrate that SMTP achieves more accurate trajectory forecasts and yields a 1.21\% IDF1 improvement over the state-of-the-art trajectory prediction module EqMotion when integrated into the same MOT framework. Overall, our SCT-MOT consistently achieves superior accuracy and robustness compared to state-of-the-art trackers across multiple metrics under complex swarm scenarios.
\end{abstract}

\begin{IEEEkeywords}
	Multiple object tracking, Swarm UAVs tracking, trajectory prediction, feature fusion.
\end{IEEEkeywords}

\section{INTRODUCTION}

\begin{figure}[!t]
	\centering
	\subfloat[]{\includegraphics[width=\linewidth]{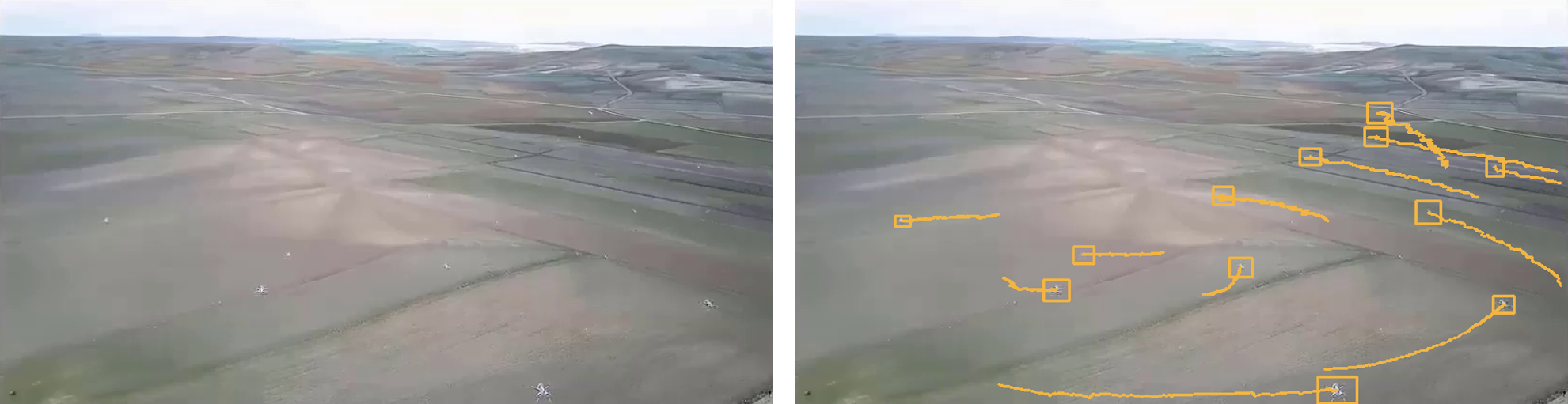}%
		\label{subfig: uavswarm-15}}
	\vspace{0mm} 
	\subfloat[]{\includegraphics[width=\linewidth]{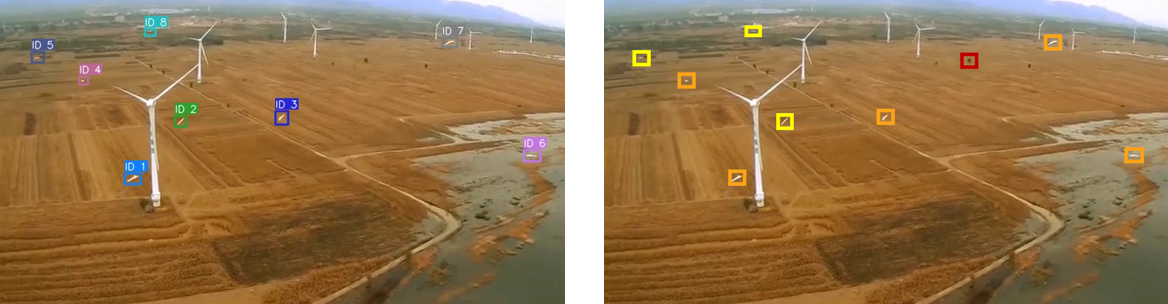}%
		\label{subfig: uavswarm-24}}
	\caption{Key challenges in air-to-air swarm UAV tracking.
		(a) Swarm UAVs exhibit highly coupled and nonlinear group behavior. The left image is the raw input frame; the right shows the corresponding annotated ground-truth trajectories.
		(b) Due to extremely micro object sizes and weak visual features, tracking performance is easily affected by background noise and similar distractors, leading to issues such as misassociation, false tracking, and missed detections. The left image shows ground-truth annotations, while the right presents the tracking results generated by HOMATracker. Orange boxes denote correct tracklets, red boxes indicate false positive tracklets and yellow boxes indicate missed tracklets.}
	\label{fig_intro}
\end{figure}

A{\scshape ir}-to-air multiple object tracking (A2A-MOT) for swarm unmanned aerial vehicles (UAVs) has become increasingly important due to the rapid advancement of intelligent unmanned systems \cite{uavswarmmode, uavchallenge}. This technology supports a wide range of applications, including formation coordination \cite{swarmcommunication}, military surveillance \cite{swarmmission}, swarm perception \cite{swarmnet} and anti-drone operations \cite{dut-anti-uav}. The goal of A2A-MOT is to continuously localize and identify multiple UAVs across video frames \cite{homatracker}. Despite recent progress in multiple object tracking (MOT) \cite{MOTReview, deepsort,giaotracker,strongsort,ctracker, qdtrack, attentiontrack}, A2A-MOT in swarm scenarios remains highly challenging due to two main factors. First, swarm UAVs exhibit highly coupled and nonlinear motion patterns driven by formation constraints, which complicate motion modeling and association across frames. Second, swarm UAVs often have very small sizes and weak visual features \cite{lce-yolo}, compounded by cluttered backgrounds, makes accurate detection difficult \cite{antiUAV410}. These visual limitations cause conventional detection algorithms to produce false positives and missed detections, leading to fragmented trajectories and identity switches. As a result, maintaining consistent trajectories for UAV objects becomes particularly difficult. Fig. \ref{fig_intro} illustrates some of these challenges on the UAVSwarm dataset \cite{uavswarm} for air-to-air swarm UAV tracking. 

To mitigate detection errors and improve tracking robustness, several recent methods incorporate trajectory prediction mechanisms \cite{IMANet, ltttrack}. However, most of them model each object independently, neglecting the inter-dependencies in motion patterns caused by collective swarm behavior \cite{pid-mot}. This limitation leads to inaccurate trajectory predictions, especially in dynamic swarm scenarios. For instance, UAVs in a formation flight are strongly influenced by the position and velocity of their neighbors, and any change in one UAV’s trajectory can cause a chain reaction across the group. Ignoring these coupling effects results in poor trajectory accuracy, as the motion patterns of each UAV are not considered in relation to the others.

Moreover, even when trajectory predictions are reasonably accurate, how to effectively combine them with visual features to assist in current-frame tracking for micro UAVs remains an open challenge. Existing methods treat motion prediction as a separate process \cite{yolo-3dmm}, without deeply integrating with visual cues. Specifically, relying solely on motion cues may not sufficiently enhance the discriminative power of weak visual features, particularly for small UAV objects that are often interfered by cluttered backgrounds. In such cases, motion predictions based on historical trajectories may not align well with the UAV's current appearance features. This misalignment hinders the effective fusion of motion and appearance features, limiting the model's ability to reliably distinguish weak objects and maintain consistent tracking.

To address these limitations, we propose an air-to-air swarm UAV tracking method that incorporates swarm-coupled motion modeling and trajectory-guided feature fusion. Specifically, we first introduce a Swarm Motion-aware Trajectory Prediction (SMTP) module that jointly captures nonlinear group motion and posture-aware visual features from a swarm-level perspective, enabling precise trajectory prediction. Next, we design a Trajectory-Guided Spatio-Temporal Feature Fusion (TG-STFF) module that uses predicted trajectories and historical detection features to generate predictive feature maps. These maps are then fused deeply with current frame features to enhance temporal consistency and spatial discriminability, especially for UAV objects with weak visual cues. These fused features are subsequently utilized for downstream association and detection tasks, substantially improving performance under complex swarm motion conditions.

The main contributions of this paper are summarized as follows:

1. We propose a Swarm Motion-aware Trajectory Prediction (SMTP) module and a Trajectory-Guided Spatio-Temporal Feature Fusion (TG-STFF) module, which together enable effective swarm-coupled motion modeling and trajectory-guided visual feature fusion. 

2. We develop SCT-MOT, a dynamic feature fusion-based tracking framework tailored for air-to-air swarm UAV tracking, which significantly enhances tracking accuracy and consistency under nonlinear and collaborative group dynamics.

3. Extensive experiments on three public air-to-air swarm UAV tracking datasets-AIRMOT, MOT-FLY and UAVSwarm-demonstrate that SMTP achieves more accurate trajectory forecasts and yields a 1.21\% IDF1 improvement over the state-of-the-art trajectory prediction module EqMotion when integrated into the same MOT framework. Moreover, SCT-MOT consistently achieves superior accuracy and robustness compared to state-of-the-art trackers across multiple evaluation metrics.

\section{RELATED WORK}

\subsection{Trajectory Prediction}
Trajectory prediction aims to forecast future positions of agents based on historical observations. In multi-object tracking algorithms, classical prediction methods typically employ statistical filters such as Kalman and partical filters \cite{bytetrack,ocsort,hybridsort}, which assume predefined motion models and Gaussian noise distributions. Although computationally efficient, these methods struggle with highly nonlinear trajectories and non-Gaussian uncertainties \cite{c-bioutracker}. 

To overcome such limitations, data-driven approaches based on deep learning have gained popularity. Social-LSTM \cite{social-lstm} was among the first to incorporate social interactions into trajectory forecasting via a long short-term memory (LSTM) network and a social pooling mechanism. Subsequently, a wide range of methods have emerged, leveraging recurrent neural networks \cite{longtermrnn}, social pooling \cite{rnn}, attention mechanisms \cite{socialattention}, and graph neural networks \cite{padgcnn}. However, most of these methods rely heavily on geometric cues, such as Euclidean distance, limiting their ability to capture complex semantic context and behavioral patterns.

Recent efforts have focued on improving interaction modeling and interpretability. Tang et al. \cite{multi-agent-prediction} proposed a framework that jointly learns trajectory distributions and collaborative uncertainty. LSSTA \cite{lssta} fused graph convolutional networks (GCNs) with spatial transformers to effectively capture complex spatial-temporal dependencies. MANTRA \cite{mantra} employed an external memory module to store and adaptively refine trajectory patterns via end-to-end training. EvolveGraph \cite{EvolveGraph} dynamically constructed relational graphs to reason the evolution of inter-agent interactions. EqMotion \cite{eqmotion} further leveraged equivariant networks to model geometric consistency in multi-agent motion, thereby enabling more robust predictions. 

Although recent trajectory prediction models have achieved promising results in human crowds and traffic scenarios, they often neglect swarm-specific properties such as formation constraints, collective dynamics, and coordinated maneuvers. This limitation leads to inaccurate forecasts when facing nonlinear trajectories, synchronized motions, and formation transitions in UAV swarms, where the strong coupling among individual UAVs and collective swarm behaviors is essential for reliable prediction. To address this limitation, we propose a trajectory prediction module that explicitly captures spatial–temporal dependencies at both the individual pose level and the swarm-structure level. By integrating the global-local swarm-coupled motion patterns with appearance features, our approach enables more robust and accurate trajectory forecasting in complex air-to-air swarm tracking scenarios.

\subsection{General Multi-Object Tracking under UAV Perspectives}

Recent UAV-based MOT approaches focus on establishing reliable object associations across frames under complex motion patterns and challenging camera dynamics \cite{bactrack}. Major strategies include designing fine-grained appearance extractors, motion-aware cost functions, and adaptive feature update mechanisms. For instance, UAVMOT \cite{uavmot} introduces an adaptive strategy that correlates current detections with the top-k candidates from the previous frame to improve association robustness. MG-MOT \cite{mg-mot} addresses long-term occlusions by incorporating UAV platform metadata (e.g., altitude, pitch) into the re-identification pipeline, thereby improving association reliability. UCMCTrack \cite{ucmctrack} proposes a motion-compensated non-IoU distance metric, which projects detections from the image plane to the world coordinate system under planar-ground assumption. Deep EIoU \cite{deep-eiou} improves the tracking of non-linearly moving objects by replacing traditional Kalman filters with deep feature-guided IoU-based iterations. Meanwhile, DC-MOT \cite{dc-mot} integrates deblurring and motion compensation modules to mitigate blur-induced degradation and the effects of rapid camera motion. 
To further address false positives and missed detections, IMANet \cite{IMANet} introduces a camera-aware motion modeling module and a multi-scale detection-tracking fusion strategy. However, while effective in general UAV tracking scenarios, its performance degrades in swarm UAV tracking due to the micro object size, low visual saliency and frequent nonlinear group motions. Its fusion strategy lacks the ability to effectively integrate weak appearance features, which are common in densely packed UAV swarms.

Different from previous work, our trajectory-guided spatio-temporal feature fusion module incorporates both historical detection features and swarm-consistent spatial priors. By explicitly aligning visual features with predicted trajectories, our approach enhances temporal consistency and spatial discriminability, significantly reducing false associations and improving tracking stability in visually ambiguous and dynamic swarm scenes.

\subsection{Air-to-Air Swarm UAVs Tracking}
Research on air-to-air swarm UAVs tracking remains relatively limited, primarily due to the scarcity of large-scale benchmark datasets. Existing public datasets, such as MOT-FLY \cite{MOT-FLY}, UAVSwarm \cite{uavswarm} and AIRMOT \cite{homatracker}, have recently emerged to facilitate research in this domain, enabling the evaluation of tracking algorithms under challenging aerial conditions. Most existing methods focus on enhancing visual discriminability and improving association robustness. For instance, UAVS-MOT \cite{uavs-mot} extends FairMOT \cite{fairmot} by integrating a coordinate attention module to boost appearance-based discrimination for swarm UAV objects. BELGTracker \cite{belgtracker} introduces a cascaded multi-UAV tracking framework that leverages local geometric constraints and morphology-aware feature enhancement to improve identification accuracy. HOMATracker \cite{homatracker} proposes a multi-frame pose-attention mechanism for UAV appearance modeling, along with a motion-difference accumulation strategy to capture nonlinear UAV movement over time. While these approaches demonstrate promising results, they often suffer from weak visual features that lead to detection failures (false positives and missed detections) and lack robustness when handling the complex, coupled motion patterns inherent in swarm behavior. These limitations lead to frequent association errors and ID switches, particularly in scenarios involving dense formations or coordinated maneuvers. 

In this work, we propose a dynamic feature fusion framework specifically designed for swarm UAV tracking in air-to-air scenarios. Unlike previous methods that treat appearance-based discrimination and motion modeling separately, our approach integrates swarm-coupled motion cues through the Swarm Motion-Aware Trajectory Prediction (SMTP) module, which models the nonlinear motion of UAVs from a swarm-level perspective. Additionally, the Trajectory-Guided Spatio-Temporal Feature Fusion (TG-STFF) module fuses predictive motion features with historical detection information, enhancing weak visual features by improving the spatio-temporal consistency of UAV objects. This dual fusion mechanism improves both the temporal coherence and spatial discriminability of weak objects, significantly boosting tracking accuracy and robustness in scenarios involving multi-formations and nonlinear, coordinated UAV maneuvers. 

\begin{figure*}[htp]
	\centering
	\includegraphics[width=1.0\linewidth]{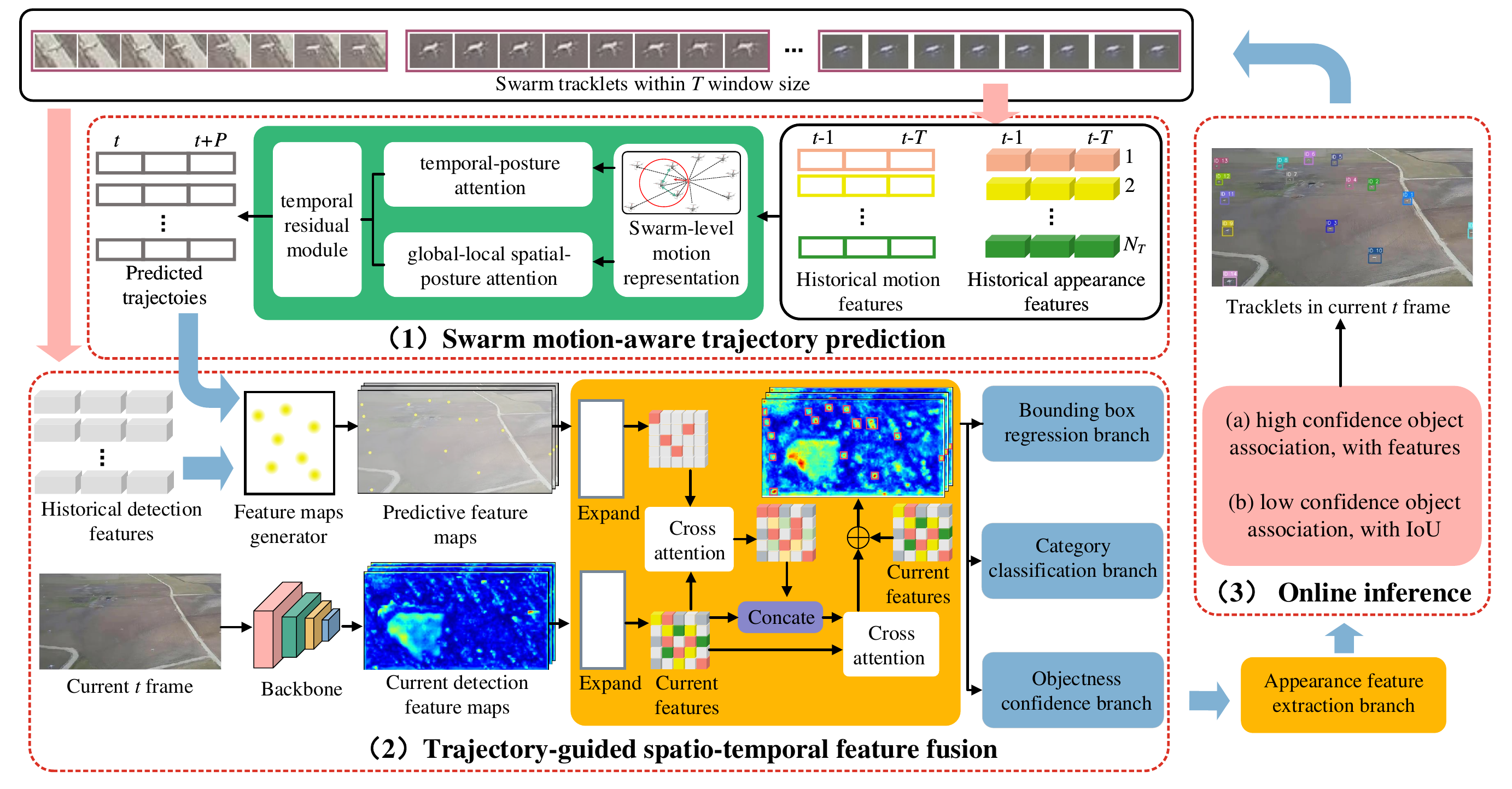}
	\caption{{The overall architecture of the SCT-MOT framework. The swarm motion aware trajectory prediction, trajectory-guided spatio-temporal feature fusion, detection and tracking after trajectory-guided fusion and online inference are introduced in Sec. \ref{subsection: SMTP}, Sec. \ref{subsection: TG-STFF}, Sec. \ref{subsection: detection and tracking branch}, and Sec. \ref{subsection: association}, respectively.} }
	\label{fig_1}
\end{figure*}

\section{METHOD}

This section presents SCT-MOT, a multi-object tracking framework specifically designed to address the challenges of air-to-air swarm UAV tracking. We first propose an overview of our SCT-MOT architecture in subsection A. Subsequently, subsection B and C detail two core components: the Swarm Motion-aware Trajectory Prediction (SMTP) module and the Trajectory-Guided Spatio-Temporal Feature Fusion (TG-STFF) module. Subsection D describes the detection and tracking heads along with the corresponding training loss functions. Finally, the online inference procedure is explained in subsection E.

\subsection{Architecture Overview} \label{subsection: overview}
The SCT-MOT framework follows the tracking-by-detection paradigm, where object detection and tracking are performed in separate stages. Notably, the proposed Swarm Motion-aware Trajectory Prediction (SMTP) and Trajectory-Guided Spatio-Temporal Feature Fusion (TG-STFF) modules are designed as plug-and-play components, enabling seamless integration into both tracking-by-detection and joint detection–tracking frameworks.

As illustrated in Fig. \ref{fig_1}, the overall architecture of SCT-MOT comprises three main components: 1) A Swarm Motion-aware Trajectory Prediction (SMTP) module, which predicts current positions of swarm UAVs by jointly modeling their historical motion and appearance features. 2) A Trajectory-Guided Spatio-Temporal Feature Fusing (TG-STFF) module, which generates predictive feature maps and fuses them with current frame features to enhance spatial-temporal representations for subsequent detection and tracking. 3) An online inference pipeline that associates UAVs detected via fused features with their corresponding tracklet identities. 

For a typical tracking-by-detection pipeline or an existing single-stage tracker, the proposed SMTP module can be incorporated into the back-end of the tracker, following the data association stage, to refine tracklet predictions based on swarm-level temporal cues. Meanwhile, the TG-STFF module can be integrated into the backbone feature extraction stage as an auxiliary feature fusion branch, enhancing the spatio-temporal representation of UAV objects without altering the original detection head, thereby ensuring compatibility and ease of integration with existing trackers.

During the tracking process, we define a swarm tracklet as an object that consistently appear across all frames within a $T$ temporal window. Given a video sequence from frames $I_{t-T}$ to $I_{t-1}$, the historical tracking features of swarm tracklet $i$ is defined as $sw_i=\{\{e_{t-T,i},c_{t-T,i}\},\cdots,\{e_{t-1,i},c_{t-1,i}\}\}$, where $e_{{t_j},i} \in \mathbb{R}^{d}$ denotes the appearance embedding produced by the tracker, and $c_{t_j,i} \in \mathbb{R}^{n}$ the spatial position at frame $t_j$ ($n=2$ for 2D image space). The feature set of all $N_T$ active swarm tracklets is denoted as $SW_{N_T}=\{sw_1,\cdots,sw_{N_T}\}$. These features are processed by the SMTP module to predict their current positions at frame $t$: 
$C^{pred}_t=\{c^{pred}_{t,1},c^{pred}_{t,2},\cdots,c^{pred}_{t,N_T}\}$. Meanwhile, the current frame $I_t$ is encoded by the backbone into a set of multi-scale detection feature maps $\mathcal{F}_t=\{F^1_t,\cdots,F^m_t\}$, where each $F^m_t\in \mathbb{R}^{H_l\times W_l\times C_l}$ corresponds to a specific downsampling level $l\in\{8,16,32\}$, with $H_l$ and $W_l$ denoting the spatial resolution and $C_l$ the channel dimension. The historical detection features within $T$ temporal window are defined as $F_D = \{\{f_{t-T,i}, \cdots, f_{t-1,i}\}, \cdots, \{f_{t-T,N_T}, \cdots, f_{t-1,N_T}\}\}$, where each $f_{t_j,i}$ is obtained by applying ROI max pooling over the bounding box region of UAV $i$ in $\mathcal{F}_{t_j}$. By jointly utilizing $C^{pred}_t$, $\mathcal{F}_t$, and $F_D$, the TG-STFF module generates the fused feature maps $\mathcal{M}_t$. $\mathcal{M}_t$ is then fed into the detection and tracking branches to produce the final tracking results $\mathcal{T}_t=\{(c_{t,i},id_i, e_{t,i}, s_{t,i})\}^{N_T}_{i=1}$ for the current $t$ frame, where $c_{t,i}$, $id_i$, $e_{t,i}$ and $s_{t,i}$ denote the position, identity, appearance feature and confidence score of UAV $i$, respectively.   

\subsection{Swarm Motion-aware Trajectory Prediction} \label{subsection: SMTP}
We propose a trajectory prediction module that leverages the spatial-temporal coupling between individual UAV pose and collective swarm motion. In swarm UAV scenarios, a UAV's visual appearance inherently reflects its motion posture. For example, a forward-leaning UAV tends to move in the corresponding direction in subsequent frames. This implicit pose information is embedded in the appearance features extracted from each UAV, enabling us to model motion trends without explicitly estimating pose. Additionally, swarm UAVs exhibit collective motion patterns, where the trajectory of each UAV is influenced by its neighbors through dynamic interactions such as obstacle avoidance and swarm following. At the group level, swarm objects often exhibit coordinated motion trends and maintain specific geometric formations. Our SMTP module jointly models the spatio-temporal dependencies between individual pose cues and group-level motion dynamics to accurately predict the future positions of swarm UAVs.

\begin{figure*}[htp]
\centering
\includegraphics[width=1.0\linewidth]{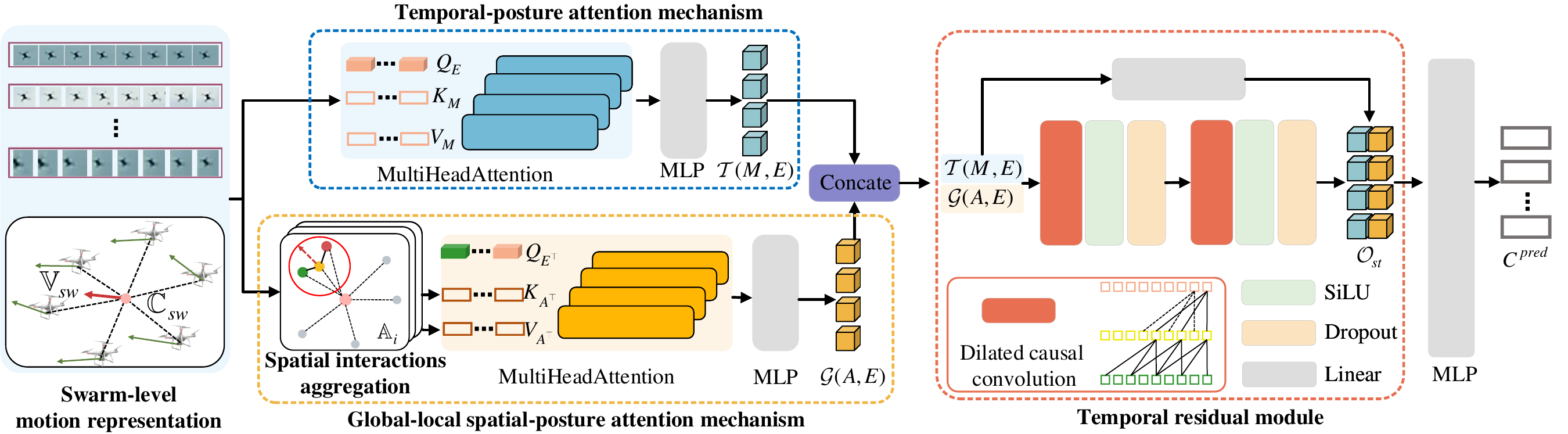}
\caption{{The overall architecture of the SMTP module. This module includes: a temporal-posture attention mechanism, a global-local spatial-posture attention mechanism, and a temporal residual module. These submodules jointly capture posture-aware temporal dynamics, spatial-aware posture representations and integrate these features for trajectory prediction.} }
\label{fig_2}
\end{figure*}

As shown in Fig. \ref{fig_2}, for each of the $N_T$ swarm UAVs observed across a temporal window of $T$ frames, we construct a set of spatio-temporal features comprising position, velocity and appearance embedding. Specifically, the historical positions are organized into tensor $C_{sw}=\{C_1,\cdots,C_{N_T}\} \in \mathbb{R}^{N_T \times T \times n}$. Each $C_i = \{c_{t-T,i}, \cdots, c_{t-1,i}\}$ denotes the 2D image-plane center coordinates $(x,y)$ of UAV $i$ over the past $T$ frames.
Similarly, the velocity tensor is defined as $V_{sw}=\{V_1,\cdots,V_{N_T}\} \in \mathbb{R}^{N_T \times T \times n}$. Each $V_i= \{v_{t-T,i},\cdots, v_{t-1,i}\}$ represents the historical velocities of UAV $i$, where $v_{t_j,i}=c_{t_j,i}-c_{t_{j-1},i}$. The corresponding appearance features form the tensor $E_{sw}=\{E_1,\cdots,E_{N_T}\} \in \mathbb{R}^{N_T \times T \times d}$. 

To characterize global swarm motion, we compute the frame-wise mean position and velocity across all swarm tracklets as: 
\begin{equation}
\begin{aligned}
	\mathbb{C}_{sw}=\{\frac{1}{N_T}\sum^{N_T}_{i=1}c_{t_j,i}\}^{t-1}_{t_j=t-T} \in \mathbb{R}^{T\times n}, \\
	\mathbb{V}_{sw}=\{\frac{1}{N_T}\sum^{N_T}_{i=1}v_{t_j,i}\}^{t-1}_{t_j=t-T} \in \mathbb{R}^{T\times n}
\end{aligned}
\label{eq:1}
\end{equation}
where the mean operation is computed across all $N_T$ UAVs at each frame $t_j$. These swarm-level statistics reflect the evolving collective motion trend and approximate formation center of the swarm. 

We then compute the relative motion features for each individual UAV by subtracting the swarm-level statistics and projecting both individual and swarm features into a shared $f$-dimensional latent space ($f=128$ in our implementation) for subsequent spatio-temporal modeling:
\begin{equation}
\begin{aligned}
	\mathbb{C}_i &= W_{c_i}(C_i - \mathbb{C}_{sw}) + W_{c, sw} \mathbb{C}_{sw} \in \mathbb{R}^{T \times f}, \\
	\mathbb{V}_i &= W_{v_i}(V_i - \mathbb{V}_{sw}) + W_{v, sw} \mathbb{V}_{sw} \in \mathbb{R}^{T \times f}, \\
	\mathbb{E}_i &= W_{e_i}(E_i) \in \mathbb{R}^{T \times f}
\end{aligned}
\label{eq:sw}
\end{equation}
where $W_{c_i}, W_{v_i}, W_{e_i}, W_{c,sw},$ and $W_{v,sw} \in \mathbb{R}^{n\times f}$ are learnable linear projection matrices. This formulation enables each UAV to encode both individual motion deviations and global swarm trends.

The motion representation $\mathbb{M}_i$ for UAV $i$ is then obtained by concatenating its encoded position and velocity features along the feature dimension, and mapping the resulting tensor through a learnable fusion layer $W_{fuse} \in \mathbb{R}^{2f\times f}$:
\begin{equation}
\begin{aligned}
	\mathbb{M}_i=W_{fuse}([\mathbb{C}_i;\mathbb{V}_i])\in \mathbb{R}^{T\times f}, \quad  {M}=\{\mathbb{M}_1,\cdots,\mathbb{M}_{N_T}\}
\end{aligned}
\end{equation}
where $[\cdot \, ; \cdot]$ denotes concatenation along the feature dimension.

To model the temporal relationship between UAV motion and appearance, we introduce a multi-head temporal-posture attention mechanism with $k = 8$ heads, which allows the model to focus on different aspects of the motion-appearance interaction across multiple attention heads. Specifically, the queries are derived from the temporal appearance features ${E}=\{\mathbb{E}_1,\cdots,\mathbb{E}_{N_T}\}\in \mathbb{R}^{N_T\times T\times f}$, while the keys and values are derived from the motion features $M$. The aggregated attention process is formulated as:
\begin{equation}
	\begin{aligned}
		Q^j_{E}&=W^j_{Q} {E},\quad K^j_{M}=W^j_{K} {M},\quad V^j_{M}=W^j_{V} M \\
		\text{head}_j&=\text{softmax}\left(\frac{Q^j_{E}(K^j_{M})^\top}{\sqrt{d_k}}\right)V^j_{M}, \quad j=1,\dots,k\\
		\mathcal{T}(M,E)&=\text{MLP}_t\left( \text{Concat}(\text{head}_1,\cdots, \text{head}_k) \right)\in \mathbb{R}^{N_T\times T\times f}
	\end{aligned}
\end{equation}
where $d_k = f / k = 16$, and $W^j_Q$, $W^j_K$, and $W^j_V \in \mathbb{R}^{f\times d_k}$ are learnable linear projection matrices for the $j$-th head. The attention outputs from all heads are concatenated and processed by a two-layer MLP with SiLU activation to obtain the final posture-aware temporal dynamics $\mathcal{T}(M,E)$.

Building upon the temporal modeling and global swarm-level motion representations, we further incorporate local spatial coupling among neighboring UAVs. For each swarm tracklet $sw_i$, we define its neighborhood as the set of its $ne$ nearest UAVs located within a spatial radius $L$, and $ne$ is set to 2 in our implementation. The spatial interaction between UAV $i$ and its neighboring UAV $j$ is computed by fusing their velocity embeddings and relative positional encodings:
\begin{equation}
	A_{ij}=W_{ne\_fuse}([\mathbb{V}_i;\mathbb{V}_j;W_{neighbor}(\mathbb{C}_i-\mathbb{C}_j)])\in \mathbb{R}^{T\times f}
	\label{eq:5}
\end{equation}
where $W_{ne\_fuse} \in \mathbb{R}^{3f\times f}$ and $W_{neighbor} \in \mathbb{R}^{f\times f}$ are linear layers, and $(\mathbb{C}_i - \mathbb{C}_j)$ represents the the relative position trajectory across $T$ frames.

The local interaction representation $\mathbb{A}_i$ for UAV $i$ is then constructed by aggregating the spatial interactions from its neighbors, which is subsequently fused with its own velocity information:
\begin{equation}
	\mathbb{A}_i=\mathbb{C}_i+W_{a_i}([\mathbb{V}_i;\sum_{j\in \text{Neighbor}(i)}A_{ij}])\in \mathbb{R}^{T\times f}
\end{equation}
where the linear layer $W{a_i}\in R^{2f\times f}$, and the result is added to the original position embedding $\mathbb{C}_i$ via residual connection.

To capture both global posture semantics and local spatial dependencies, we propose a global-local spatial-posture attention mechanism. Specifically, the local interaction features $\{ \mathbb{A}_1, \cdots, \mathbb{A}_{N_T} \}$ are reshaped into $A^\top \in \mathbb{R}^{T \times N_T \times f}$ and serve as the keys and values, while the appearance features $E^\top \in \mathbb{R}^{T \times N_T \times f}$ are used as queries.

The global-local spatial-posture attention $\mathcal{G}(A,E)$ is formulated as a standard multi-head scaled dot-product attention:
\begin{equation}
	\mathcal{G}(A,E)=\text{MultiHeadAtt}(Q_{E^\top},K_{A^\top},V_{A^\top})\in \mathbb{R}^{T\times N_T\times f}
\end{equation}
where $Q_{E^\top} = W^j_Q E^\top$, $K_{A^\top} = W^j_K A^\top$, and $V_{A^\top} = W^j_V A^\top$, $j=1,\dots,k$ are the projected queries, keys, and values respectively. The outputs of all $k=8$ heads are concatenated to obtain the spatial-aware posture representation of each UAV.

We concatenate the posture-aware temporal dynamics and spatial-aware posture posture representations to form the input sequence $\mathcal{S}$, which is then processed by a temporal residual module consisting of two layers of dilated causal convolutions. Specifically:
\begin{equation}
	\begin{aligned}
		&\mathcal{S}=\text{Concat}(\mathcal{T}(M,E),\mathcal{G}(A,E)^\top), \\
		&\mathcal{U}(\mathcal{S},i,t)=\sum^{K-1}_{k=0}W_k\cdot \mathcal{S}_{t-k\cdot r}(i)\in \mathbb{R}^{N_T\times T\times d_{out}}, \\
		&\mathcal{S}_{t-k\cdot r}(i)=\mathcal{S}[i;t-k\cdot r,:]
	\end{aligned}
\end{equation}
where $\mathcal{S}_{t-k\cdot r}(i)$ represents the feature of UAV $i$ at the $(t-k\cdot r)$-th frame, sampled with a dilation rate $r$ along the temporal axis to enlarge the receptive field. The causal convolution ensures that only past frames are used for prediction. $W_k\in \mathbb{R}^{2f\times d_{out}}$ are learnable weights, $r=2$ is the dilation factor, $K=3$ is the kernel size and $d_{out}$ is the output dimension. 

The final spatio-temporal posrue representation $\mathcal{O}_{st}$ is calculated by combining the residual and convolutional outputs:
\begin{equation}
	\begin{aligned}
		\mathcal{O}_{st}=W'_k \mathcal{S}+\mathcal{U}(\mathcal{S}), S_{st}\in \mathbb{R}^{N\times T\times d_{out}}
	\end{aligned}
\end{equation}	
where $W'_k\in \mathbb{R}^{2f\times d_{out}}$ is a learnable projection matrix. 

This fused representation is then extended along the temporal axis and passed through a two-layer MLP with SiLU activation to predict the future positions of UAVs over the next $P$ frames:
\begin{equation}
	C^{pred}=\text{SiLU}(\mathcal{O}_{st}W_0)W_1\in \mathbb{R}^{N_T\times P\times n}
\end{equation}
where $W_0$ and $W_1$ are the learnable decoding weights. 

The entire module is trained by minimizing the $L^2$ loss between the predicted and ground-truth future positions:
\begin{equation}
	\mathcal{L}_\text{motion}= \frac{1}{N_TP}\sum^{N_T}_{i=1}\sum^{P}_{p=1}||C^{pred}_{p,i}-C^{gt}_{p,i}||_2
\end{equation}
where $C^{gt} \in \mathbb{R}^{N_T \times P \times n}$ denotes the ground-truth future positions, and the loss is averaged across all swarm tracklets and prediction steps.

\subsection{Trajectory-Guided Spatio-Temporal Feature Fusion} \label{subsection: TG-STFF}
To address the challenge of detecting and tracking the high-dynamic and extremely small UAVs, we introduce a Trajectory-Guided Spatio-Temporal Feature Fusion (TG-STFF) module. It incorporates prior cues from two complementary sources: (1) historical visual features extracted from detected UAV tracklets, and (2) trajectory predictions of swarm tracklets predicted by the SMTP module. These priors are used to guide the current detection and tracking process, thereby enhancing robustness against missed and false detections. 

\begin{figure}[t]
	\centering
	\includegraphics[width=1.0\linewidth]{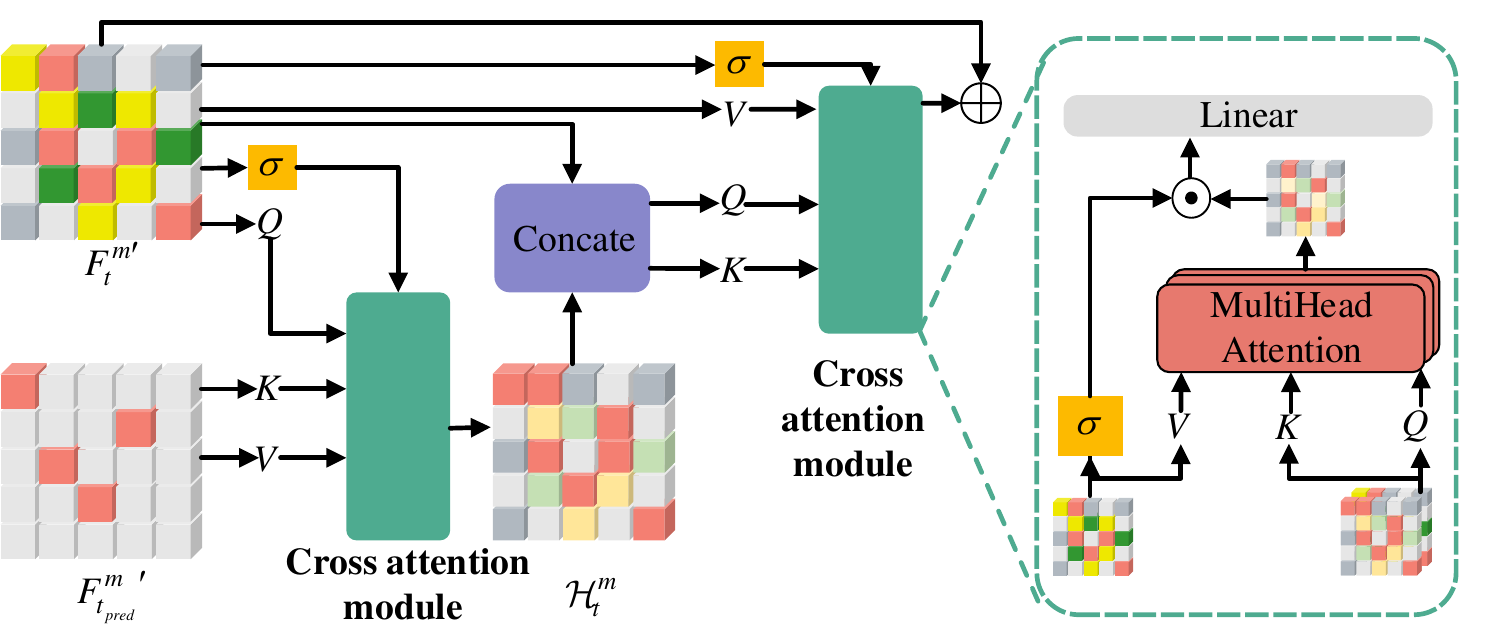}
	\caption{{The architecture of the TG-STFF module. ${F^{m}_t}'\in \mathbb{R}^{D_l\times C_l}$ represents the expanded feature map of the current frame, and ${F^m}_{t_{pred}}\in \mathbb{R}^{D_l\times C_l}$ denotes the Gaussian-distributed predicted feature map. We use cross-attention module to fuse them to generate the final spatio-temporal feature map.} }
	\label{fig_3}
\end{figure}

As shown in Fig. \ref{fig_3}, the current frame $I_t$ is encoded by the detection backbone to produce multi-scale feature maps $\mathcal{F}_t$. For the previous $T$ frames $I_{t-T}$ to $I_{t-1}$, we extract object-level detection features for each tracked UAV and organize them as $F_D=\{\mathbb{F}_1,\cdots, \mathbb{F}_{N_T}\}\in \mathbb{R}^{N_T\times T\times f}$, where $\mathbb{F}_i=\{f_{t-T,i},\cdots,f_{t-1,i}\}$ denotes the historical visual feature of UAV $i$. Each feature $f_{t_j,i}$ is extracted from the detection feature map $\mathcal{F}_{t_j}$ by applying spatial max pooling over the region corresponding to the historical bounding box $B_i$. 
:
\begin{equation}
	\begin{aligned}
		f_{t_j,i}=\mathcal{P}_{\text{max}}(\mathcal{F}_{t_j}(B_i))
	\end{aligned}
\end{equation}
where $\mathcal{F}_{t_j}(B_i)$ represents the feature region corresponding to the bounding box $B_i$, which is extracted via RoI-Align, and $\mathcal{P}_{\text{max}}$ denotes spatial max pooling operation to yield an $f$-dimensional vector.

Let $C^{pred}_t=\{c^{pred}_{t,1},c^{pred}_{t,2},\cdots,c^{pred}_{t,N_T}\}$ denote the predicted locations of all $N_T$ swarm tracklets at frame $t$. To guide the fusion process, we utilize three sources of information: (1) the current multi-scale feature maps $\mathcal{F}_t$, (2) the predicted locations $C^{pred}_t$, and (3) the historical visual features $F_D$ for swarm tracklets. 

We first construct a predictive feature map ${F^m_{t_{pred}}}$ using $C^{pred}_t$ and $F_D$, where each UAV’s historical visual embedding $\mathbb{F}_i$ is projected into the feature space and distributed over a Gaussian kernel centered at its predicted location:
\begin{equation}
	{F^{m}_{t_{pred}}}(:,:,x,y)=\sum^{N_T}_{i=1}W_{cat}(\mathbb{F}_i)\cdot exp(-\frac{||(x,y)-C^{pred}_t(i)||^2}{2\sigma^2})
\end{equation}
where $W_{cat}\in \mathbb{R}^{f\times C_l}$ projects the concatenated historical features $\mathbb{F}_i\in \mathbb{R}^{T\times f}$ into the channel dimension. 

Both $F^m_t\in\mathcal{F}_t$ and the generated ${F^{m}_{t_{pred}}}$ are reshaped into ${F^m_t}', {F^{m}_{t_{pred}}}' \in \mathbb{R}^{D_l\times C_l}$, where $D_l=H_l\times W_l$ and $C_l$ is the channel dimension at scale $m$. A multi-head cross-attention mechanism is then used to compute the spatial-temporal correlation $\mathcal{H}^m_t\in \mathbb{R}^{D_l\times C_l}$ between the two:
\begin{equation}
	\mathcal{H}^m_t=(\sigma({F^m_t}'W^t_1)\ast \text{MultiHeadAtt}({F^m_t}',{F^{m}_{t_{pred}}}',{F^{m}_{t_{pred}}}'))W^o_1
\end{equation}
where $\text{MultiHeadAtt}$ denotes a standard multi-head scaled dot-product attention module, computing correlations between the current-frame features (${F^m_t}'$ as queries) and the predictive features (${F^{m}_{t_{pred}}}'$ as keys and values). The element-wise multiplication $\ast$ applies channel-wise modulation using $\sigma({F^m_t}'W^t_1)$ as a spatial attention gate. The number of attention heads is set to 4, and $\sigma$ denotes the activation function.

To further enrich the spatial-temporal representation, we concatenate $\mathcal{H}^m_t$ with the original ${F^m_t}'$ and apply a second cross-attention module to obtain the final enhanced feature ${F^m_t}''\in\mathbb{R}^{D_l\times C_l}$:
\begin{equation}
	\begin{aligned}
		&{F^m_t}''= {F^m_t}'+(\sigma({F^m_t}'W^t_2)\ast\\
		&\text{MultiHeadAtt}([{F^m_t}' ; \mathcal{H}^m_t],[{F^m_t}' ; \mathcal{H}^m_t],{F^m_t}'))W^o_2
	\end{aligned}
\end{equation}
where $[\cdot\, ;\, \cdot]$ denotes channel-wise concatenation and $\mathcal{H}^m_t$ provides the context-enhanced spatial-temporal features. $W^t_2$, $W^o_2 \in \mathbb{R}^{C_l \times C_l}$ are learnable projection matrices.

Finally, the enhanced feature ${F^m_t}''$ is reshaped back to the original spatial dimensions $(H_l, W_l, C_l)$, forming ${F^m_{t_{fuse}}}\in \mathbb{R}^{H_l\times W_l \times C_l}$, enabling seamless integration into the detection feature hierarchy. Across all scales $m$, the final fused multi-scale spatio-temporal feature maps are collected as:
\begin{equation}
	\mathcal{M}_t = \{ F^1_{t_{fuse}}, \cdots, F^m_{t_{fuse}} \}
\end{equation}

Since the fused features preserve the same spatial and channel dimensions as the original detection feature maps, TG-STFF module can be directly integrated into existing tracking frameworks without requiring additional loss functions.

\subsection{Detection and Tracking Branch after Trajectory-Guided Fusion} \label{subsection: detection and tracking branch}
Based on the dynamically fused multi-scale spatio-temporal feature maps produced by the TG-STFF module, we design the detection and tracking branches that jointly perform four types of predictions: bounding box regression, category classification, objectness confidence, and appearance features. 

The bounding box regression branch aims to predict the locations of objects' bounding boxes. We denote the $i$-th predicted bounding box as $b_i=\{x^{pred}_i,y^{pred}_i,w^{pred}_i,h^{pred}_i\}$, and the corresponding ground truth as $\hat{b}_i=\{\hat{x}_i,\hat{y}_i,\hat{w}_i,\hat{h}_i\}$, the regression loss is defined as:
\begin{equation}
	\mathcal{L}_\text{reg}=-\frac{1}{N}\sum_{i=1}^N\text{log}(\text{IoU}(\hat{b}_i, b_i))
\end{equation}
where $\text{IoU}(\cdot,\cdot)$ denotes the intersection over union, and $N$ is the total number of ground truth boxes. This loss penalizes inaccurate localization and encourages higher overlap during training.

The category classification loss is computed using cross-entropy between predicted class probabilities and the ground truth labels. It is formulated as: 
\begin{equation}
	\mathcal{L}_\text{cls}=-\frac{1}{N}\sum_{i=1}^{N}\sum^M_{c=1}y_{ic}\text{log}(p_{ic})
\end{equation}
where $M$ is the number of object categories, and $y_{ic}\in \{0,1\}$ is a binary indicator that equals 1 if the $i$-th instance belongs to class $c$, and 0 otherwise.

%For objectness prediction branch, the confidence loss is defined as:
The objectness branch predicts whether a proposal corresponds to valid UAV objects. The binary confidence loss is computed as:
\begin{equation}
	\mathcal{L}_\text{obj}=-\frac{1}{N_{obj}}\sum^{N_{obj}}_i[\hat{p}_i\text{log}(p_i)+(1-\hat{p}_i)\text{log}(1-p_i)]
\end{equation}
where $p_i$ is the predicted objectness probability at location $i$, $\hat{p}_i\in \{0,1\}$ is the corresponding ground truth label, and $N_{obj}$ denotes the number of candidate locations.

To support identity association across frames, we adopt the appearance feature extraction branch from HOMATracker \cite{homatracker}, a state-of-the-art framework for swarm UAV tracking. This module is based on a multi-frame pose attention mechanism, which captures spatially localized appearance cues guided by UAV motion and posture information. Concretely, for each tracklet, this module generates a set of part-level appearance features $\{E'_1, E'_2, \dots, E_n'\}$, by aggregating visual features across multiple frames via a posture attention mechanism. These part features are then concatenated to form the full appearance features $E$ for association. To ensure efficiency, the appearance extraction branch is implemented using the lightweight ResNet18 backbone. The corresponding association loss is defined to encourage high similarity between features of the same UAV across frames and low similarity between different UAVs. It is formulated as:
\begin{equation}
	\begin{aligned}
		\mathcal{L}_{j_\text{asso}}(E'_j, \tau_m)=-\sum_{t_j=1}^{T'}\sum_{i\in\mathcal{D}_{t_j}}\text{log}P_A(i,\tau_m|t_j)
	\end{aligned}
\end{equation}

\begin{equation}
	\mathcal{L}_\text{asso}(E,\mathcal{T})=\sum_{\tau_m\in\mathcal{T}}\sum_{j=1}^n\mathcal{L}_{j_\text{asso}}(E'_j,\tau_m)
\end{equation}
where $\mathcal{D}_{t_j}$ represents the set of detections in frame $t_j$, $\tau_m$ denotes the $m$-th tracklet, and $P_A(i,\tau_m|t_j)$ represents the similarity between detection $i$ and tracklet $\tau_m$ at frame $t_j$, $\mathcal{T}$ denotes the set of active tracklets. The parameter $T'$ specifies the temporal association window size, and $n$ is the number of part-level feature embeddings extracted from each UAV's tracklet.

The overall optimization is performed in multiple stages. We first pretrain the detector using $\mathcal{L}_\text{det}$. 
\begin{equation}
	\mathcal{L}_\text{det}=\lambda\cdot\mathcal{L}_\text{reg}+\mathcal{L}_\text{obj}+\mathcal{L}_\text{cls}
\end{equation}
The weighting factor $\lambda=5$ is empirically chosen to balance the scale differences between the regression and classification losses.
Subsequently, with detector weights initialized from the pretrained model, we train the motion and association branches under $\mathcal{L}_\text{assoc}$.
\begin{equation}
	\mathcal{L}_\text{assoc}=\mathcal{L}_\text{motion}+\mathcal{L}_\text{asso}
\end{equation}

\subsection{Online Inference} \label{subsection: association}
For the online tracking stage, we adopt a multi-frame association framework tailored for swarm UAV scenarios, leveraging both motion and appearance cues within a sliding temporal window. The inference process operates in real time, dynamically updating tracklet identities as new frames arrive. 

In the first frame, high confidence detections are initialized as individual tracklets. Subsequently, a sliding temporal window of length $T'$ is employed for online inference. At the current frame $t$, the set of tracklets within the sliding window is denoted as $\mathcal{T}=\{\tau_1,\dots\tau_K\}$, where $K$ is the number of tracklets in the window. The detections in the current frame $I^t$ are divided into high confidence detections $\mathcal{D}_t^{high}=\{d^t_i,\dots,d^t_{N_t}\}$ and low confidence detections $\mathcal{D}_t^{low}=\{d^t_{N_t+1},\dots,d^t_{M}\}$, where $N_t$ denotes the number of high-confidence detections and $M$ is the total number of detections in frame $t$.

For high confidence detections, we extract pose-appearance features $E\in \mathbb{R}^{N_t\times d}$ using the proposed feature extraction branch and compute the similarity matrix $M_A$ between these detections and the existing tracklets $\mathcal{T}$. Meanwhile, we compute the spatial motion similarity matrix $M_S$ using the distance cost calculation strategy from HOMATracker, which measures the normalized spatial displacement between current detections and and their corresponding historical positions across frames within the temporal window. These two cues are integrated via element-wise multiplication to produce the final association cost matrix:
\begin{equation}
	\mathbf{M}_F=M_A\otimes M_S
\end{equation}
Specifically, $M_A, M_S \in \mathbb{R}^{K \times N_t}$, where $K$ is the number of existing tracklets in $\mathcal{T}$.

We apply the Hungarian algorithm on $\mathbf{M}_F$ to associate detections with tracklets. Unmatched high confidence detections are then initialized as new tracklets. For low confidence detections $\mathcal{D}_t^{low}$, we perform short-term association based on the IoU between detections and tracklets in the previous frame $t-1$. Only matched detections are retained, while unmatched ones are discarded.

\section{EXPERIMENTS}

\subsection{Datasets and Evaluation Metrics}
To evaluate the performance of our proposed SCT-MOT, we conduct experiments on three publicly available swarm UAV tracking datasets: AIRMOT, UAVSwarm and MOT-FLY. These datasets encompass diverse tracking scenarios, including homogeneous UAV tracking, heterogeneous UAV tracking, and swarm UAV tracking under dynamic conditions.

AIRMOT is a simulated dataset for air-to-air homogeneous swarm UAV tracking. It contains 8 RGB video sequences comprising 7,844 frames (5,124 for training and 2,720 for testing), and a resolution of $1920\times 1080$. Each frame includes 5-16 UAV instances of the same type, generated in the AirSim simulator under varying lighting, backgrounds, and camera view angles. The UAVs exhibit various formations and complex motion patterns,  resulting in nonlinear trajectories in the 2D image plane.

MOT-FLY is a real-world UAV tracking dataset consisting of 16 RGB sequences (11,186 frames total), with 7,238 frames for training and 3,948 for testing. Each sequence contains 1-3 UAV instances of different types. Over $90\%$ of UAVs occupy less than $5\%$ of the image area. This dataset captures a wide range of visual conditions and motion dynamics, posing significant challenges for tracking evaluations.

UAVSwarm is an open-source benchmark for swarm-level UAV tracking, comprising 72 sequences across 13 distinct scenarios and more than 19 UAV models. It provides 6,844 frames in 36 sequences for training and 5,754 frames in 36 sequences for testing. The dataset features diverse camera perspectives and environmental conditions, including UAV formation transitions, rapid motion of micro UAVs, and dynamic camera movements.

We evaluate tracking performance using four standard metrics:  Multiple Object Tracking Accuracy (MOTA), Identity F1 score (IDF1), Higher Order Tracking Accuracy (HOTA), and inference speed measured in Frames Per Second (FPS) \cite{MOTReview}. These metrics jointly evaluate tracking accuracy (MOTA), identity consistency (IDF1), overall spatio-temporal association quality (HOTA), and runtime efficiency (FPS).

\subsection{Implementation Details}
All experiments are conducted on an NVIDIA RTX 3090 GPU. We first pretrain the baseline tracking framework HOMATracker using input images resized to $1088\times 1088$. The model is optimized using Stochastic Gradient Descent (SGD) with an initial learning rate of 0.002, momentum of 0.9, and weight decay of 0.0005. Subsequently, we integrate our proposed SMTP and TG-STFF modules into HOMATracker. For training the SMTP module, we adopt a historical window of 8 frames to predict future positions over the next 12 frames. In each frame, object appearance features and historical trajectories are extracted and fed into this module. The predicted positions are supervised using ground truth trajectories throught the position prediction loss. Notably, our TG-STFF module does not require an additional loss function and is trained end-to-end along with the entire network. 

During inference, the historical window size $T$ is set to 8 with a stride of 1. Starting from the 9th frame, the SMTP module predicts the current positions of swarm objects using their past trajectories. These predicted positions and historical detection features are used to construct trajectory-guided predictive feature maps, which are then fused with current features via the TG-STFF module. The final fused feature maps are then fed into detection and tracking branches. 
We further adopt a sliding window of size $T'=16$ and step size 1 for online tracking. All input frames are resized to $1088\times 1088$ while maintaining a fixed aspect ratio. Detections are filtered using a confidence threshold of 0.01, and Non-Maximum Suppression (NMS) is applied with an IoU threshold of 0.1. Tracklet association is performed using motion and appearance cues, while unmatched high-confidence detections are initialized as new tracklets.
\begin{figure}[!t]
	\centering
	\subfloat[]{\includegraphics[width=\linewidth]{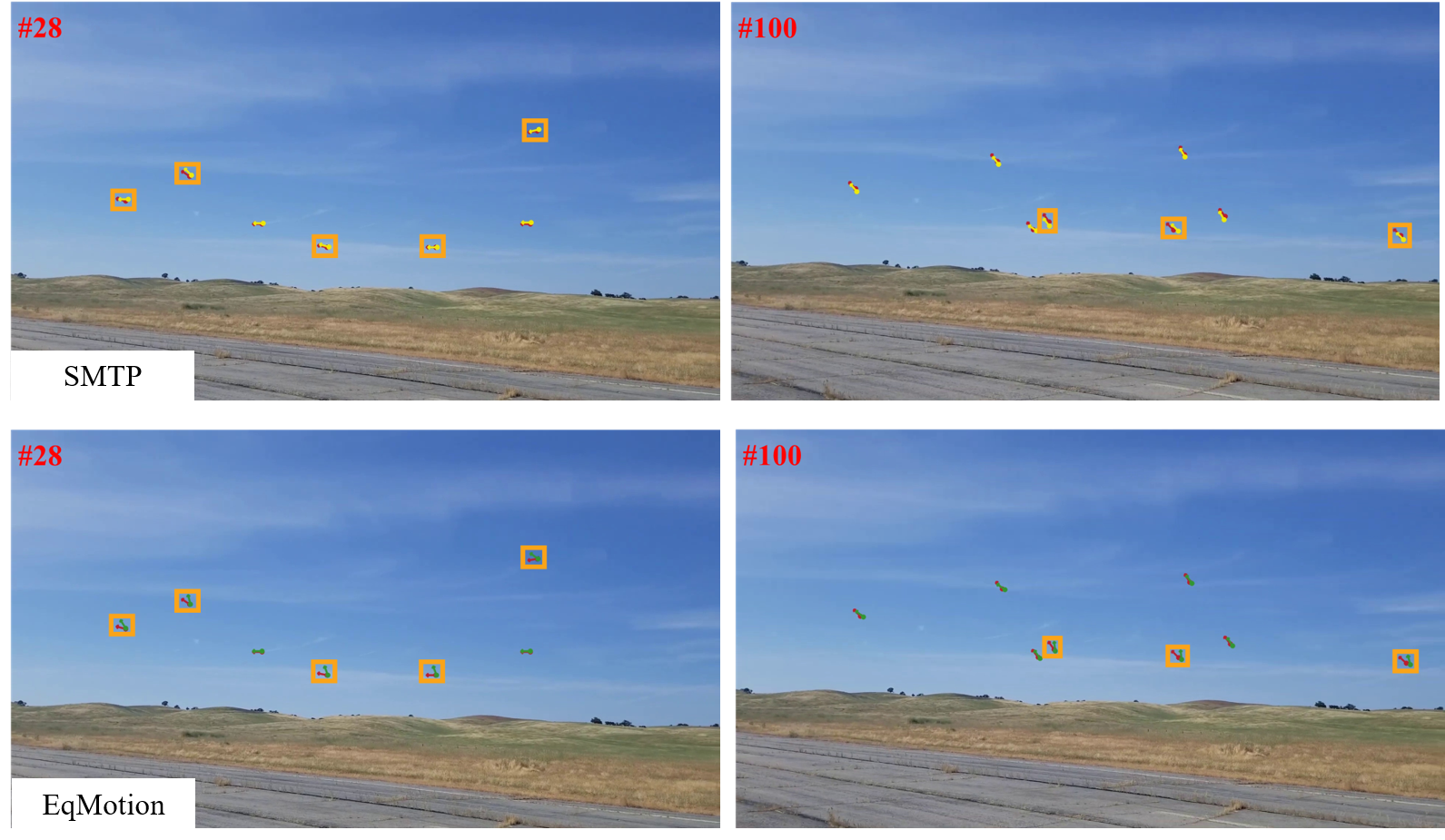}%
		\label{subfig: uavswarm-40}}
	\vspace{0mm} 
	\subfloat[]{\includegraphics[width=\linewidth]{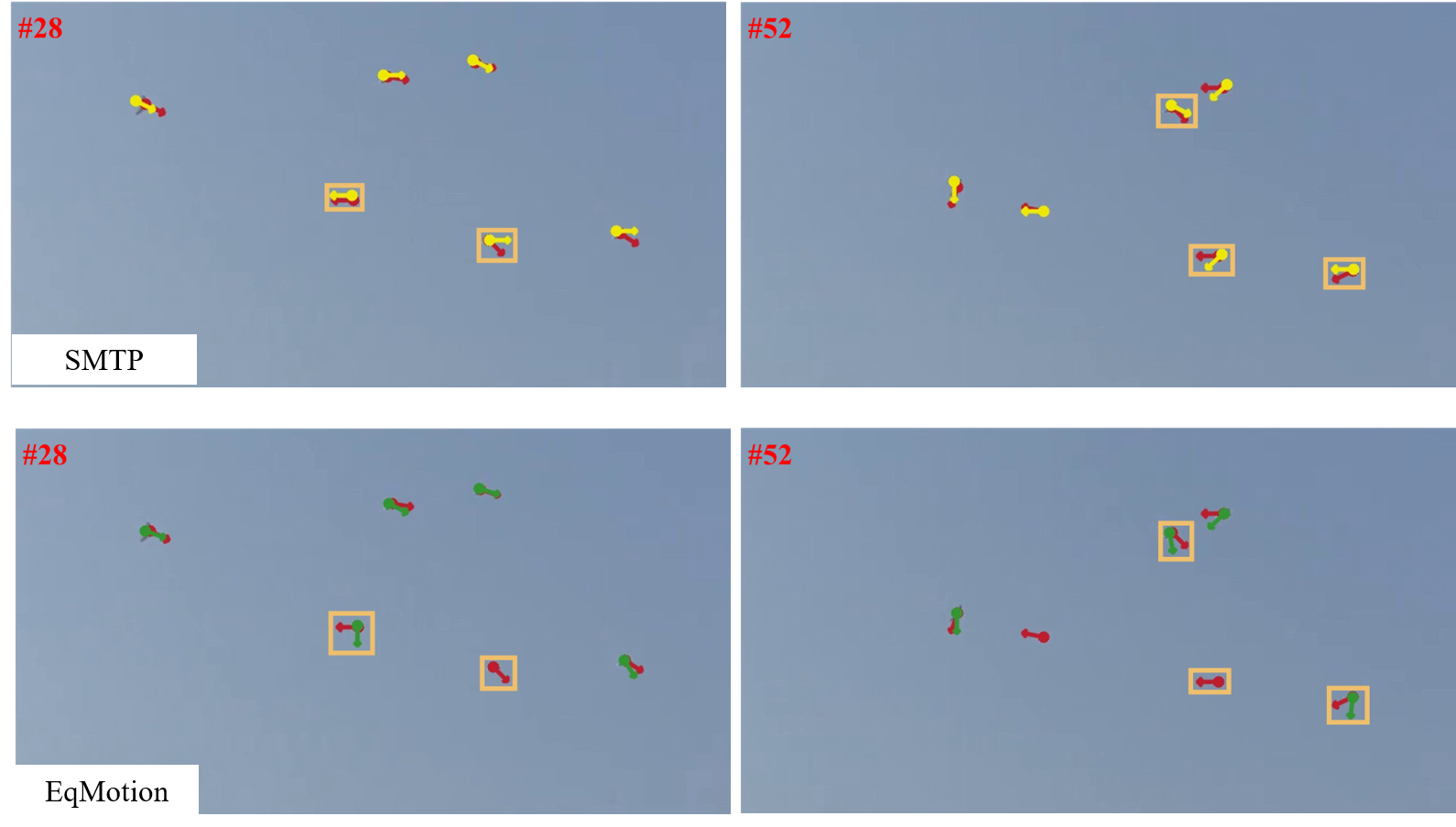}%
		\label{subfig: uavswarm-58}}
	\vspace{0mm} 
	\subfloat[]{\includegraphics[width=\linewidth]{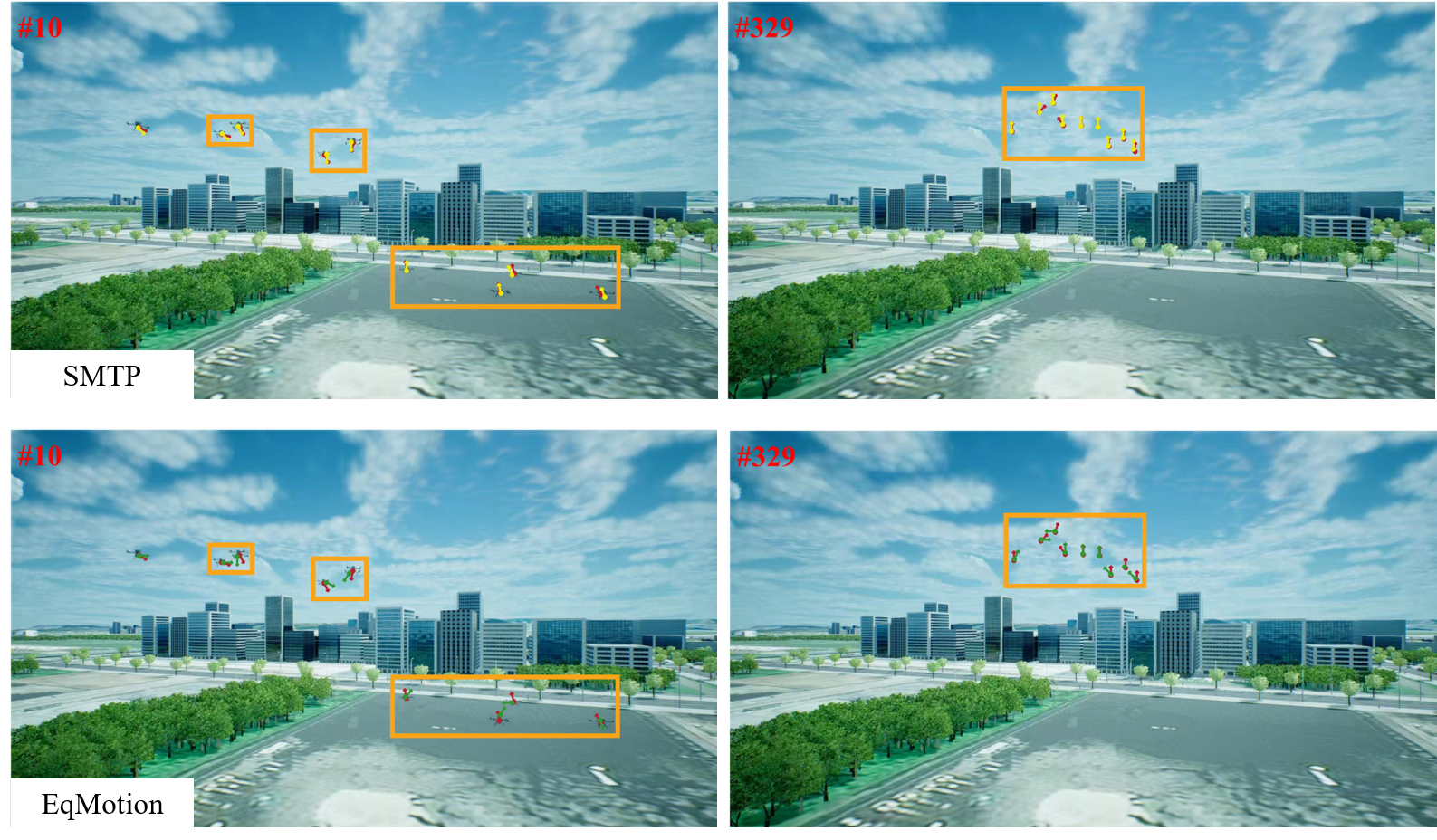}%
		\label{subfig: airmot-M300}}
	\caption{Trajectory prediction comparison between different modules in air-to-air swarm tracking. Red dots and arrows denote the ground-truth positions and motion directions; yellow dots and arrows indicate the predicted positions and directions by the proposed SMTP module; green ones represent the predictions by EqMotion. Orange boxes highlight areas where SMTP predictions are more closely aligned with the ground-truth trajectories. The subfigures illustrate representative sequences from the UAVSwarm and AIRMOT datasets: (a) UAVSwarm-40, (b) UAVSwarm-58, and (c) AIRMOT-01.}
	\label{fig_arrow}
\end{figure}

\subsection{Ablation Studies}
\subsubsection{Swarm Motion-aware Trajectory Prediction}
Our SMTP module jointly captures posture-aware visual cues and historical trajectories from a swarm-level perspective, enabling accurate prediction of spatially coupled UAV motion patterns. We first investigate the impact of the historical window size $T$ (set to 4, 8, 12, and 16) on the performance of this module using the AIRMOT and UAVSwarm datasets. SMTP is integrated into the SCT-MOT framework, the resulting multi-object tracking metrics are shown in Table \ref{tab_smtp_T}.

On AIRMOT, $T$ = 4 achieves the highest MOTA (33.63\%) and the lowest IDSW (457), while both HOTA and IDF1 remain close to optimal. This suggests that in low-coupling scenarios where swarm UAVs move with less inter-object dependency, a short windows is sufficient to maintain trajectory continuity and reduce ID switches. Longer windows ($T$ = 12, 16) result in degraded performance, likely due to the accumulation of outdated or less-relevant historical information, which increases the risk of prediction errors.
In comparison, the UAVSwarm dataset exhibits more consistent and spatially coupled cluster motion. A moderate window ($T$ = 8) yields the best results across all metrics (MOTA: 81.90\%, IDF1: 88.45\%, HOTA: 68.56\%) and achieves the lowest IDSW (56), demonstrating the effectiveness of a moderate window length in dense, formation-based scenarios. When $T$ increases further ($T$ = 12, 16), prediction accuracy declines, potentially due to cumulative prediction error or occlusions in the extended trajectory. The differences in optimal window size across datasets may due to varying levels of coupling between posture changes and motion patterns. In the AIRMOT dataset, UAVs perform significant posture changes when transitioning between motion states, making a shorter window effective for capturing these rapid dynamics. In contrast, UAVs in the UAVSwarm dataset maintain minimal posture variations, making a moderate window size more suitable for capturing inter-agent motion dependencies and local dynamics.

Building upon the analysis of the historical window size, we further evaluate the overall effectiveness of SMTP by quantitatively comparing it with the state-of-the-art trajectory prediction method EqMotion \cite{eqmotion} module on both datasets. For a fair comparison, we replace the trajectory prediction component in the SCT-MOT framework with either SMTP or EqMotion, while keeping all other network modules identical. The resulting tracking performance is summarized in Table \ref{tab_smtp}. As shown in the table, integrating SMTP yields consistently better tracking performance compared to EqMotion. On AIRMOT, SMTP improves MOTA, IDF1, and HOTA by 0.39\%, 1.21\%, and 0.62\%, respectively, while reducing ID switches. On UAVSwarm, it achieves improvements of 1.18\%, 0.88\%, and 0.87\% on the same metrics, alongside fewer ID switches. These results confirm the superior trajectory prediction capability of SMTP in both sparse and dense swarm scenarios.

In addition to the quantitative results, we also perform a visual comparison to further illustrate how SMTP improves trajectory prediction quality over EqMotion in complex swarm scenarios.
Fig. \ref{fig_arrow} illustrates the representative prediction results under three challenging swarm tracking scenarios from both UAVSwarm and AIRMOT datasets:
(a) irregular camera motion with frequent UAVs entry and exit the field of view;
(b) highly dynamic swarm maneuvers;
(c) complex multi-formation transitions. 
As shown in Fig. \ref{fig_arrow}, SMTP consistently produces more accurate predictions in both spatial positions and motion directions across various swarm motion patterns, while EqMotion exhibits larger deviations in complex group behaviors. For example, in Fig. \ref{fig_arrow} (c), when faced with complex multi-formation transitions, the predicted UAV positions and velocity directions by EqMotion deviate significantly from the ground truth. This suggests that EqMotion fails to account for the collective motion constraints imposed by the swarm during formation transitions, which are crucial for accurately predicting individual UAV behavior in such scenarios. In contrast, SMTP’s predictions are closely aligned with the ground truth, demonstrating that by incorporating a global swarm-level perspective, SMTP more effectively models the influence of group-level motion patterns on individual UAVs, especially in scenarios where UAVs exhibit complex, strongly coupled movements. These results highlight the superior capability of SMTP in modeling motion consistency and spatio-temporal interactions of swarm UAVs, thereby mitigating deviations caused by multi-formation transitions, dynamic swarm maneuvers and viewpoint changes. 

\begin{table}
	\centering
	\caption{Evaluation of the SCT-MOT with SMTP module on AIRMOT and UAVSwarm datasets. $\uparrow$ indicates higher is better, $\downarrow$ indicates lower is better. Best results are highlighted in \textbf{bold}.}
	\label{tab_smtp_T}
	\begin{tabularx}{\linewidth}{p{0.8cm}		
			>{\centering\arraybackslash}X 
			>{\centering\arraybackslash}X 
			>{\centering\arraybackslash}X 
			>{\centering\arraybackslash}X}
%		\toprule
		\toprule
		$T$  & MOTA$\uparrow$ & IDF1$\uparrow$ & HOTA$\uparrow$ & IDSW$\downarrow$ \\
		\toprule
		AIRMOT & & & & \\
		\midrule
		4  & \textbf{33.63} & 30.04 & 25.51 & \textbf{457} \\
		8 & 32.30 & \textbf{30.15} & \textbf{25.54} & 476 \\
		12 & 31.89 & 29.61 & 25.47 & 498 \\
		16 & 31.12 & 28.85 & 25.38 & 513 \\
		%\bottomrule
		\midrule
		UAVSwarm & & & & \\
		\midrule
		4  & 81.86 & 88.15 & 68.33 & 59 \\
		8 & \textbf{81.90} & \textbf{88.45} & \textbf{68.56} & \textbf{56} \\
		12 & 81.28 & 88.02 & 67.86 & 66 \\
		16 & 80.99 & 87.30 & 67.25 & 61 \\
		\bottomrule
%		\bottomrule
	\end{tabularx}
\end{table}

\begin{table}
	\centering
	\caption{Comparison of different trajectory prediction modules for Multi-Object Tracking on AIRMOT and UAVSwarm test sets.}
	\label{tab_smtp}
	\begin{tabularx}{\linewidth}{p{2.8cm}		
			>{\centering\arraybackslash}X 
			>{\centering\arraybackslash}X 
			>{\centering\arraybackslash}X 
			>{\centering\arraybackslash}X}
%		\toprule
		\toprule
		algorithms  & MOTA$\uparrow$ & IDF1$\uparrow$ & HOTA$\uparrow$ & IDSW$\downarrow$ \\
		\toprule
		AIRMOT & & & & \\
		\midrule
		SCT-MOT-SMTP  & \textbf{32.30} & \textbf{30.15} & \textbf{25.54} & \textbf{476} \\
		SCT-MOT-EqMotion & 31.91 & 28.94 &24.92 & 478\\
		%\bottomrule
		\midrule
		UAVSwarm & & & & \\
		\midrule
		SCT-MOT-SMTP  & \textbf{81.90} & \textbf{88.45} & \textbf{68.56} & \textbf{56} \\
		SCT-MOT-EqMotion  & 80.95 & 87.08 & 67.42 & 73  \\
		\bottomrule
%		\bottomrule
	\end{tabularx}
\end{table}

\subsubsection{Trajectory-Guided Spatio-Temporal Feature Fusion}
\begin{figure*}[!ht]
	\centering
	\includegraphics[width=1.0\linewidth]{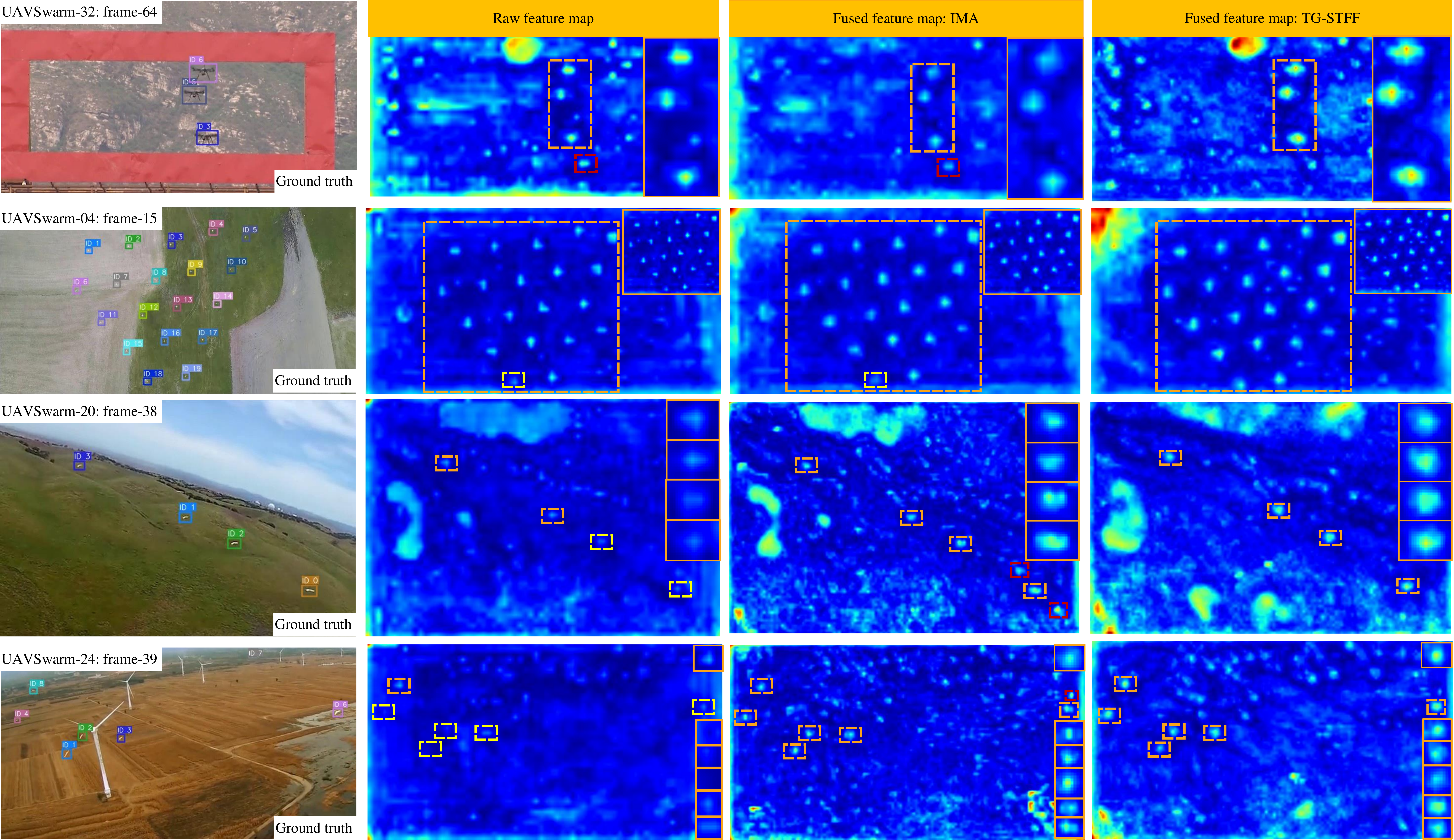}
	\caption{{Visualization of feature fusion in the TG-STFF module. From left to right: ground-truth frame, raw feature map, fused feature map using the IMA strategy, and fused feature map using TG-STFF module. Orange boxes: correct detections; red: false positives; yellow: missed detections.} }
	\label{fig_heat_map}
\end{figure*}

To validate the effectiveness of the proposed TG-STFF module in enhancing swarm-level feature representation, we conduct qualitative analysis on several UAVSwarm video sequences to visualize the effects of different fusion strategies. As shown in Fig. \ref{fig_heat_map}, we compare three types of feature maps: the raw feature map of the current frame, the feature map fused using the Interactively Motion-Assisted (IMA) strategy from IMANet \cite{IMANet}, and the feature map fused using our TG-STFF module. All fusion strategies are guided by predicted trajectories generated by the SMTP module. The IMA strategy, originally proposed in IMANet, is a representative feature fusion method leveraging interactive motion information, and serves as a baseline for comparison.

In sequences such as UAVSwarm-04, 20, 24 and 32, the raw feature maps show sparse or ambiguous activations in the object regions, leading to frequent missed and false detections. The IMA fusion strategy moderately enhances feature responses by leveraging motion-assisted cues, but struggles in scenarios with blurred or weak appearance cues. For instance, in the UAVSwarm-20 sequence, although the object region's features in frame 38 are enhanced compared to the raw feature maps, false detections occur due to the misalignment between the motion features and the current frame's detection features. In UAVSwarm-04, while feature enhancement is observed, missed detections still occur. In contrast, after applying the TG-STFF module, the object region’s features are consistently enhanced without any false detections. This improvement is attributed to TG-STFF’s dual-guided fusion design: by explicitly aligning current-frame features with historical cues along predicted trajectories, TG-STFF captures both temporal continuity and motion-aware spatial priors. Furthermore, the integration of multi-frame context allows the module to compensate for weak or missing visual evidence in any single frame, thus significantly improving tracking accuracy in weak feature scenarios.

\begin{figure*}[htp]
	\centering
	\subfloat[AIRMOT-01 sequence]{\includegraphics[width=1.0\linewidth]{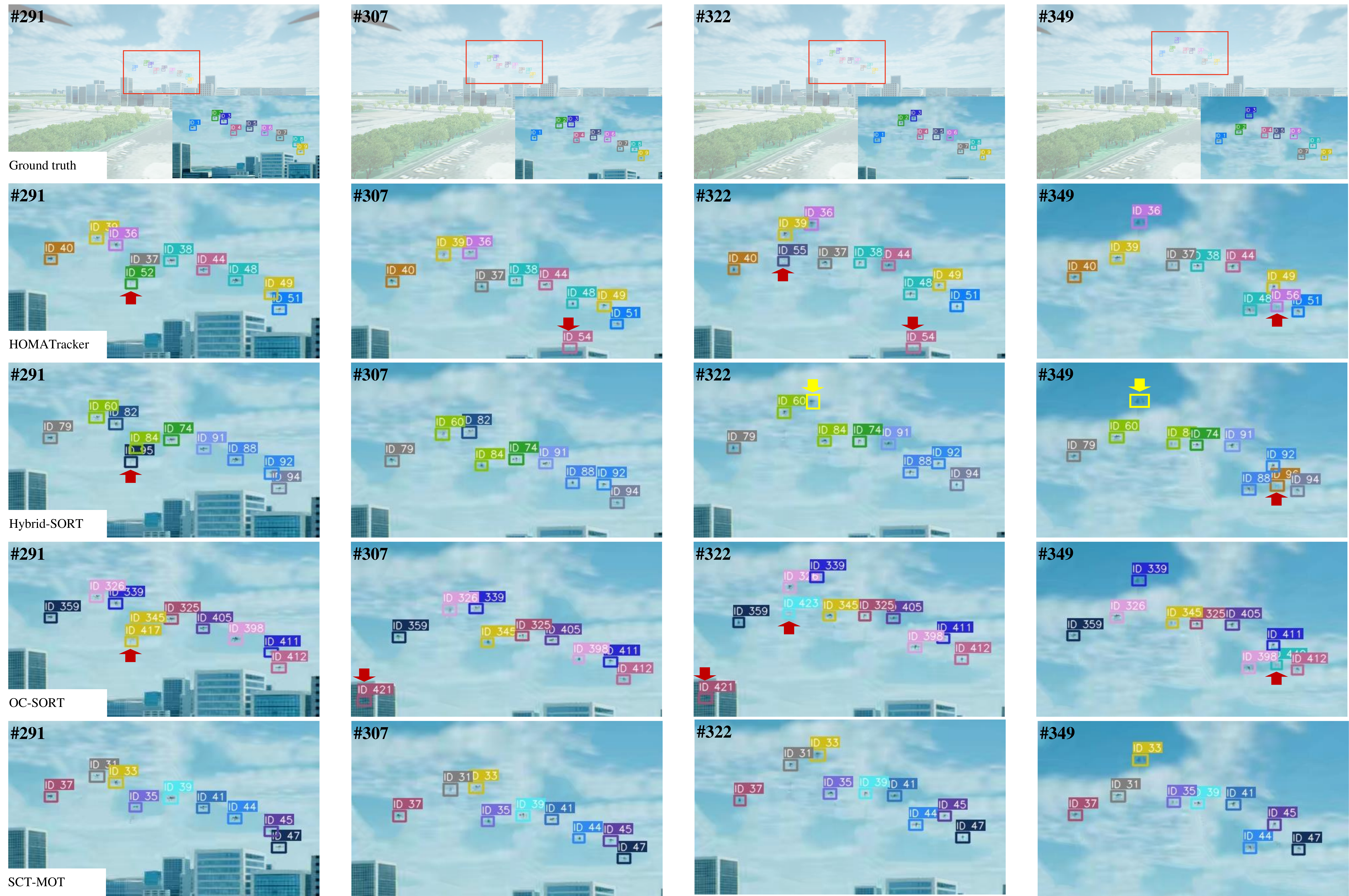}%
		\label{subfig: airmot}}
	\vspace{0mm} 
	\subfloat[UAVSwarm-16 sequence]{\includegraphics[width=1.0\linewidth]{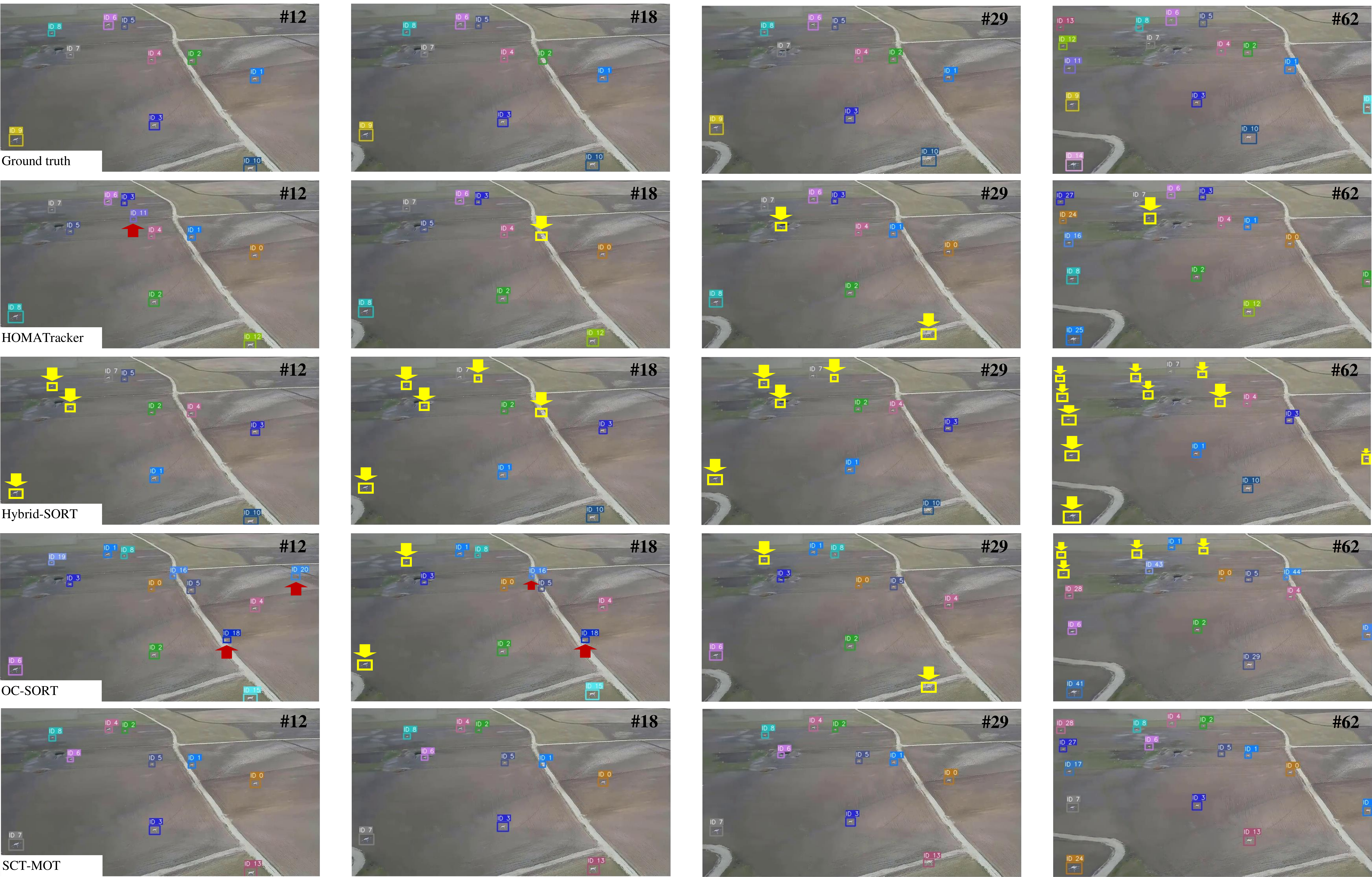}%
		\label{subfig: uavswarm}}
	\caption{Performance comparison of SCT-MOT with existing MOT methods on AIRMOT and UAVSwarm datasets, different objects are bounding boxes with different colors. Red boxes/arrows denote false positives, while yellow boxes/arrows indicate missed detections.}
	\label{fig_perform}
\end{figure*}

We also perform quantitative evaluations by integrating TG-STFF into two representative MOT frameworks: HOMATracker (tailored for UAV swarms) and FairMOT (a general MOT framework). Both trackers are integrated with the same trajectory prediction module (SMTP), enabling a fair comparison between the IMA and TG-STFF fusion strategies under unified setup. As shown in Table \ref{tab:ablation_airmot}, \ref{tab:ablation_mot-fly} and \ref{tab:ablation_uavswarm}, TG-STFF consistently improves tracking performance across all datasets and framworks. For example, on the AIRMOT dataset, TG-STFF boosts FairMOT’s MOTA, IDF1, and HOTA by 0.67\%, 1.80\%, and 1.03\%, respectively, while reducing ID switches by 101. For HOMATracker, the corresponding improvements are 1.17\%, 1.73\%, and 0.71\%, respectively. Similar improvements are observed on the MOT-FLY and UAVSwarm datasets, where TG-STFF outperforms the IMA strategy across all metrics.

\begin{table}
	\centering
	\caption{Performance comparison of FairMOT and HOMATracker using different feature fusion strategies on the AIRMOT test set.}
	\label{tab:ablation_airmot}
	\begin{tabular}{l|p{0.2cm}c|p{0.6cm}p{0.45cm}p{0.55cm}clclclc}
		\toprule
		Algorithms & {IMA} & {TG-STFF} & {MOTA}$\uparrow$ & {IDF1}$\uparrow$ & {HOTA}$\uparrow$ & {IDSW}$\downarrow$ \\
		\midrule
		FairMOT & & & 18.19 & 17.41 & 17.56 & 476 \\
		FairMOT & $\checkmark$ & & 21.19 & 19.78 & 19.38 & 516 \\
		FairMOT & & $\checkmark$ & \textbf{21.86} & \textbf{21.58} & \textbf{20.41} & \textbf{415} \\
		\midrule
		HOMATracker & & & 30.08 & 29.31 & 25.55 & 506 \\
		HOMATracker & $\checkmark$ & & 31.13 & 28.42 & 24.83 & 483 \\
		HOMATracker & & $\checkmark$ & \textbf{32.30} & \textbf{30.15} & \textbf{25.54} & \textbf{476} \\
		\bottomrule
	\end{tabular}
\end{table}

\begin{table}
	\centering
	\caption{Performance comparison of FairMOT and HOMATracker using different feature fusion strategies on the MOT-FLY test set.}
	\label{tab:ablation_mot-fly}
	\begin{tabular}{l|p{0.2cm}c|p{0.6cm}p{0.45cm}p{0.55cm}clclclc}
		\toprule
		Method & IMA & TG-STFF & MOTA$\uparrow$ & IDF1$\uparrow$ & HOTA$\uparrow$ & IDSW$\downarrow$ \\
		\midrule
		FairMOT &  &  & \textbf{63.52} & 54.81 & 42.10 & 177 \\
		FairMOT & \checkmark &  & 62.13 & 61.90 & 47.68 & 184 \\
		FairMOT &  & \checkmark & 62.96 & \textbf{63.52} & \textbf{48.32} & \textbf{172} \\
		\midrule
		HOMATracker &  &  & 72.42 & 77.64 & 58.38 & 20 \\
		HOMATracker & \checkmark &  & 72.46 & 77.90 & 58.53 & 18 \\
		HOMATracker &  & \checkmark & \textbf{72.52} & \textbf{79.31} & \textbf{59.02} & \textbf{18} \\
		\bottomrule
	\end{tabular}
\end{table}

\begin{table}
	\centering
	\caption{Performance comparison of FairMOT and HOMATracker using different feature fusion strategies on the UAVSwarm test set.}
	\label{tab:ablation_uavswarm}
	\begin{tabular}{l|p{0.2cm}c|p{0.6cm}p{0.45cm}p{0.55cm}clclclc}
		\toprule
		Method & IMA & TG-STFF & MOTA$\uparrow$ & IDF1$\uparrow$ & HOTA$\uparrow$ & IDSW$\downarrow$ \\
		\midrule
		FairMOT &  &  & 67.70 & 73.20 & 59.20 & 590 \\
		FairMOT & \checkmark &  & 69.66 & 76.40 & 59.64 & 381 \\
		FairMOT &  & \checkmark & \textbf{71.29} & \textbf{77.50} & \textbf{60.55} & \textbf{372} \\
		\midrule
		HOMATracker &  &  & 79.20 & 87.10 & 67.00 & 58 \\
		HOMATracker & \checkmark &  & 79.43 & 87.76 & 67.31 & \textbf{49} \\
		HOMATracker &  & \checkmark & \textbf{81.90} & \textbf{88.45} & \textbf{68.56} & 56 \\
		\bottomrule
	\end{tabular}
\end{table}

\begin{table*}[htp]%[!ht]
	\centering
	\caption{Performance comparison between the proposed SCT-MOT and other state-of-the-art (SOTA) MOT methods on the AIRMOT, UAVSwarm, and MOT-FLY test sets.}
	\label{tab_7}
	\begin{tabularx}{\linewidth}{p{2.5cm}l*{5}{>{\centering\arraybackslash}X}}
		\toprule
		Method & Publication & MOTA$\uparrow$ & IDF1$\uparrow$ & HOTA$\uparrow$ & IDSW$\downarrow$ & FPS$\uparrow$ \\
		\toprule
		AIRMOT & & & & & & \\
		\midrule
		FairMOT\cite{fairmot} & IJCV 2021 & 18.19 & 17.41& 17.56  & 476 & 35.2 \\
		DeepSORT\cite{deepsort} & ICIP 2017 & 19.60 & 20.45 & 19.42 & 2841 & 33.6 \\
		OCSORT\cite{ocsort} & CVPR 2023 & 20.88 & 23.27 & 22.07 & 522 & \textbf{40.9} \\
		ByteTrack\cite{bytetrack} & ECCV 2022 & 24.35 & 25.72 & 23.56 & 1707 & 39.2 \\
		MOTRv3\cite{motrv3} & ArXiv 2023 & 23.52 & 24.89 & 22.59 & 1805 & 20.6 \\
		HybridSORT\cite{hybridsort} & AAAI 2024 & 29.01 & 24.08 & 21.70 & 652 & 28.9 \\
		BELGTracker\cite{belgtracker} & \mbox{Acta Aero. Sin. (CN) 2024} & 29.27 & 23.55 & 23.47 & 1144 & 34.6 \\
		HOMATracker\cite{homatracker} & CJA 2025 & {30.08} & {29.31} & \textbf{25.55}  & {506} & {20.0} \\
		SCT-MOT & ours & \textbf{32.30} & \textbf{30.15} & {25.54} & \textbf{476} & 21.8\\
		\midrule
		MOT-FLY & & & & & \\
		\midrule
		FairMOT\cite{fairmot} & IJCV 2021 & 63.52 & 54.81 & 42.10 & 177 & 35.2 \\
		DeepSORT\cite{deepsort} & ICIP 2017 & 64.85 & 56.40 & 46.71  & 225 & 33.6 \\
		OCSORT\cite{ocsort} & CVPR 2023 & 61.41 & 66.54 & 52.25 & 194 & \textbf{40.9} \\
		ByteTrack\cite{bytetrack} & ECCV 2022 & 68.15 & 70.98 & 55.02 & 176 & 39.2 \\
		MOTRv3\cite{motrv3} & ArXiv 2023 & 65.03 & 66.74 & 52.61 & 183 & 20.6 \\
		HybridSORT\cite{hybridsort} & AAAI 2024 & 70.29 & 75.05 & 57.45 & 61 & 28.9 \\
		BELGTracker\cite{belgtracker} & \mbox{Acta Aero. Sin. (CN) 2024} & 71.92 & 73.42 & 54.85 & 118 & 34.6 \\
		HOMATracker\cite{homatracker} & CJA 2025 & {72.42} & {77.64} & {58.38} & {20} & {19.7} \\
		SCT-MOT & ours & \textbf{72.52} & \textbf{79.31} & \textbf{59.02} & \textbf{18} & 20.1 \\
		\midrule
		UAVSwarm & & & & & \\
		\midrule
		FairMOT\cite{fairmot} & IJCV 2021 & 67.7 & 73.2 & 59.2 & 590 & 35.2 \\
		DeepSORT\cite{deepsort} & ICIP 2017 & 61.2 & 70.3 & 58.8 & 221 & 33.6 \\
		OCSORT\cite{ocsort} & CVPR 2023 & 75.7 & 81.8 & 64.7 & 709 & \textbf{40.9} \\
		ByteTrack\cite{bytetrack} & ECCV 2022 & 65.0 & 76.5 & 60.6 & 67 & 39.2 \\
		MOTRv3\cite{motrv3} & ArXiv 2023 & 62.3 & 71.9 & 57.9 & 72 & 20.6 \\
		HybridSORT\cite{hybridsort} & AAAI 2024 & 77.4 & 80.1 & 62.8 & 459 & 28.9 \\
		UAVS-MOT\cite{uavs-mot} & \mbox{Acta Aero. Sin. (CN) 2024} & 73.4 & 76.1 & 65.8 & 740 & 34.6 \\
		HOMATracker\cite{homatracker} & CJA 2025 & {79.2} & {87.1} & {67.0} & {58} & {20.3} \\
		SCT-MOT & ours & \textbf{81.9} & \textbf{88.4} & \textbf{68.6} & \textbf{56} & 22.3 \\
		\bottomrule
	\end{tabularx}
\end{table*}

These results further confirm that TG-STFF enhances feature fusion by effectively leveraging predicted motion cues and spatio-temporal context, contributing to more stable and accurate tracking in dynamic UAV swarm scenarios.

\begin{table}[htp]
	\centering
	\caption{Comparison of tracking accuracy and runtime of lightweight SCT-MOT and baseline tracker on the NVIDIA Jetson Orin NX across AIRMOT, UAVSwarm, and MOT-FLY test sets.}
	\label{tab_embedded_compare}
	\begin{tabularx}{\linewidth}{p{1.8cm}		
		p{1.4cm}
		>{\centering\arraybackslash}X 
		>{\centering\arraybackslash}X 
		>{\centering\arraybackslash}X
		>{\centering\arraybackslash}X}
		\toprule
		Method & Dataset & MOTA$\uparrow$ & IDF1$\uparrow$ & HOTA$\uparrow$ & FPS$\uparrow$ \\
		\midrule
		\multirow{3}{*}{SCT-MOT} 
		& AIRMOT    & \textbf{30.68} & \textbf{28.72} & \textbf{24.31} & 24.0 \\
		& MOT-FLY   & \textbf{72.11} & \textbf{75.34} & \textbf{56.12} & 22.1 \\
		& UAVSwarm  & \textbf{78.22} & \textbf{84.04} & \textbf{66.71} & 24.5 \\
		\midrule
		\multirow{3}{*}{ByteTrack-Light} 
		& AIRMOT    & 23.46 & 24.58 & 22.86 & 24.6 \\
		& MOT-FLY   & 67.12 & 69.85 & 54.26 & 25.3 \\
		& UAVSwarm  & 63.71 & 75.13 & 59.60 & 23.9 \\
		\bottomrule
	\end{tabularx}
\end{table}

\subsection{Comparison with State-of-the-Art Methods}
To assess the performance of our proposed SCT-MOT framework in air-to-air swarm UAVs tracking scenarios, we conduct comprehensive comparisons against several representative multi-object tracking methods, including both classical models and state-of-the-art trackers: DeepSORT, ByteTrack, OC-SORT, FairMOT, MOTRv3\cite{motrv3}, Hybrid-SORT, BELGTracker, UAVS-MOT, and HOMATracker. Among them, FairMOT and UAVS-MOT follow the joint detection and tracking paradigm, MOTRv3 is state-of-the-art end-to-end transformer-based MOT method, whereas the remaining trackers adopt the tracking-by-detection paradigm. For consistency, we use YOLOX as the unified detector for all tracking-by-detection pipelines.

As shown in Table \ref{tab_7}, SCT-MOT consistently outperforms all trackers across all datasets. Notably, compared with the strong baseline HOMATracker, SCT-MOT achieves substantial improvements of +2.22\% MOTA, +0.84\% IDF1, and reduces ID switches by 30 on AIRMOT; on MOT-FLY, performance gains of +0.10\% MOTA, +1.67\% IDF1, and +0.64\% HOTA are observed. On UAVSwarm, SCT-MOT leads with +2.70\% MOTA, +1.35\% IDF1, and +1.56\% HOTA, alongside the lowest IDSW (56).

These improvements can be attributed to the integration of the robust swarm motion-aware trajectory prediction (SMTP) module and the trajectory-guided spatio-temporal fusion (TG-STFF) module, which enhances both motion forecasting accuracy and appearance feature consistency. This combination is particularly effective in air-to-air scenarios involving fast-moving, small-scale, and densely clustered UAVs, where weak visual features and nonlinear group dynamics pose significant challenges. In summary, SCT-MOT not only outperforms existing methods in tracking accuracy, but also demonstrates strong generalization across diverse scenarios and robustness to dynamic swarm behaviors, highlighting its practical potential for complex aerial tracking tasks.

We further conduct a qualitative analysis on the AIRMOT and UAVSwarm datasets. As shown in Fig. \ref{fig_perform}, we compare the inference performance of SCT-MOT with several representative trackers in scenes with substantial environmental and visual disturbances.

On the AIRMOT dataset, during frames 291 to 349, the tracking performance of existing methods significantly degrades due to environmental interference from clouds and buildings. Specifically, OC-SORT and HOMATracker exhibit repeated false positives (highlighted by red arrows), while Hybrid-SORT suffers from both missed detections and false positives (highlighted in yellow). In contrast, SCT-MOT maintains consistently high tracking accuracy throughout the sequence, successfully detecting and associating all UAVs across frames.

On the UAVSwarm dataset, which features extremely small-scale objects, dense formations, and severe background clutter, the performance differences among methods become more evident. At frame 62, Hybrid-SORT fails to detect more than half of the UAVs, while other methods also exhibit unstable detection and association. SCT-MOT, however, consistently achieves complete and robust tracking of all UAVs across the entire scene, even under severe background clutter and complex multi-formation interactions. These visual results highlight the superior robustness and swarm-level perception of SCT-MOT. By effectively modeling group dynamics and suppressing noise-induced errors, SCT-MOT achieves precise and stable tracking in complex air-to-air swarm UAV scenarios characterized by high motion complexity, limited appearance cues, and strong background interference.

\subsection{Performance on Edge Embedded Devices}
To evaluate the real-world deployability of SCT-MOT in air-to-air multi-UAV tracking, we assess its performance on a representative embedded AI platform. Considering the limited onboard computational resources in UAV, we construct a lightweight version of SCT-MOT by using HOMATracker-Light as the base framework and integrating it with the proposed SMTP and TG-STFF modules. Although these modules are not explicitly designed for lightweight inference, we employ NVIDIA TensorRT for runtime acceleration, including mixed-precision quantization and memory optimization. 

The resulting model is deployed on the NVIDIA Jetson Orin NX, a widely adopted embedded computing platform for UAV and robotics applications. We evaluate the inference performance on the AIRMOT, MOT-FLY and UAVSwarm datasets. As shown in Table \ref{tab_embedded_compare}, the lightweight SCT-MOT achieves real-time inference performance with 24.0 FPS on AIRMOT, 22.1 FPS on MOT-FLY, and 24.5 FPS on UAVSwarm. To further validate its effectiveness, we compare lightweight SCT-MOT with a representative and widely-used multi-object tracking method under the same embedded platform. We select this method as a baseline due to its balance between tracking performance and practical deployability. Notably, many state-of-the-art trackers are not originally designed for embedded inference and require extensive modification or computational resources that are not suitable for edge devices. Therefore, we focus on evaluating our algorithm against a well-recognized, practical baseline that reflects common usage scenarios in UAV edge deployments. While both methods exhibit comparable runtime performance, SCT-MOT consistently outperforms the baseline in tracking accuracy across all datasets. Specifically, it yields higher MOTA, IDF1, and HOTA scores. These results demonstrate the capability of SCT-MOT to operate efficiently and robustly on edge AI hardware, making it highly suitable for deployment in real-world embedded air-to-air UAV tracking systems.

\section{CONCLUSION}
Air-to-air tracking of swarm UAVs poses significant challenges due to the complex nonlinear group movements and weak visual cues. These factors often result in detection failures, identity switches, and fragmented trajectories, particularly under dynamic environmental interference and motion coupling. To address these challenges, we propose SCT-MOT, a multiple UAVs tracking framework that integrates swarm-aware trajectory modeling and predictive feature fusion. Specifically, The SMTP module effectively captures both posture-aware appearance cues and swarm-level motion dependencies, enabling accurate prediction of future UAV trajectories. In parallel, the TG-STFF module utilizes predicted trajectories to integrate historical and current frame information, enhancing temporal consistency and spatial discriminability. Together, these components improve detection robustness and association accuracy under complex motion and weak visual conditions. Experimental results on three public air-to-air swarm UAV tracking datasets, including AIRMOT, MOT-FLY, and UAVSwarm, show that SCT-MOT consistently achieves state-of-the-art performance across multiple metrics. This demonstrates the effectiveness of our approach in swarm UAV tracking and offers a new perspective for motion-guided feature modeling and swarm-level behavior prediction in air-to-air scenarios.

\section*{ACKNOWLEDGMENT}

This study was co-supported by the National Key Research and Development Program of China (No. 2021YFF0601304) and the National Natural Science Foundation of China (No. 62206020).

\bibliographystyle{IEEEtran}
\bibliography{IEEEtrans.bib}

\begin{IEEEbiography}[{\includegraphics[width=1in,height=1.25in,clip,keepaspectratio]{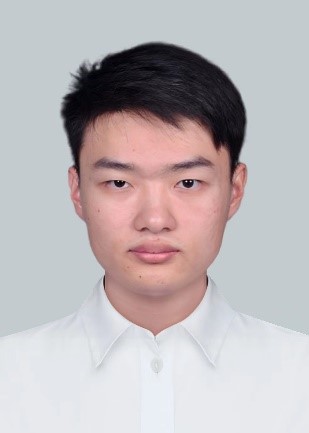}}]{Zhaochen Chu}
	received the B.E. degree in Science in Flight Vehicle Design and Engineering from Beijing Institute of Technology, Beijing, China, in 2021. He is currently pursuing the Ph.D. degree with the China-UAE Belt and Road Joint Laboratory on Intelligent Unmanned Systems of School of Aerospace Engineering, Beijing Institute of Technology. His research interests include small target UAV detection and swarm UAV tracking based on visual imagery.
\end{IEEEbiography}

\begin{IEEEbiography}
	[{\includegraphics[width=1in,height=1.25in,clip,keepaspectratio]{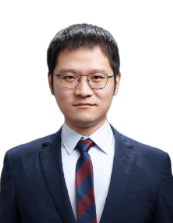}}]{Tao Song}
	received the B.E. degree in Guidance, Navigation and Control, and Ph.D. degree in Aircraft Design from Beijing Institute of Technology, Beijing, China, in 2008 and 2014, respectively. He is currently an Associate Professor with the School of Aerospace Engineering, Beijing Institute of Technology. His research interests include UAV swarm systems, intelligent aircraft modeling, and guidance and control.
\end{IEEEbiography}

\begin{IEEEbiography}
	[{\includegraphics[width=1in,height=1.25in,clip,keepaspectratio]{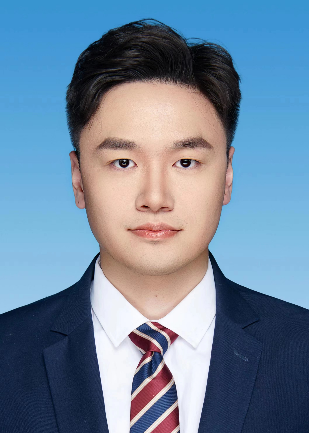}}]{Ren Jin}
	received the M.E. degree in Computer Application Technology from Hefei University of Technology, Hefei, China, in 2016, and the Ph.D. degree in Aerospace Science and Technology from Beijing Institute of Technology, Beijing, China, in 2020. He is currently a Tenure-Track Assistant Professor with the School of Aerospace Engineering, Beijing Institute of Technology. His research interests include onboard visual object detection, recognition and tracking, and UAV visual navigation.
\end{IEEEbiography}

\begin{IEEEbiography}
	[{\includegraphics[width=1in,height=1.25in,clip,keepaspectratio]{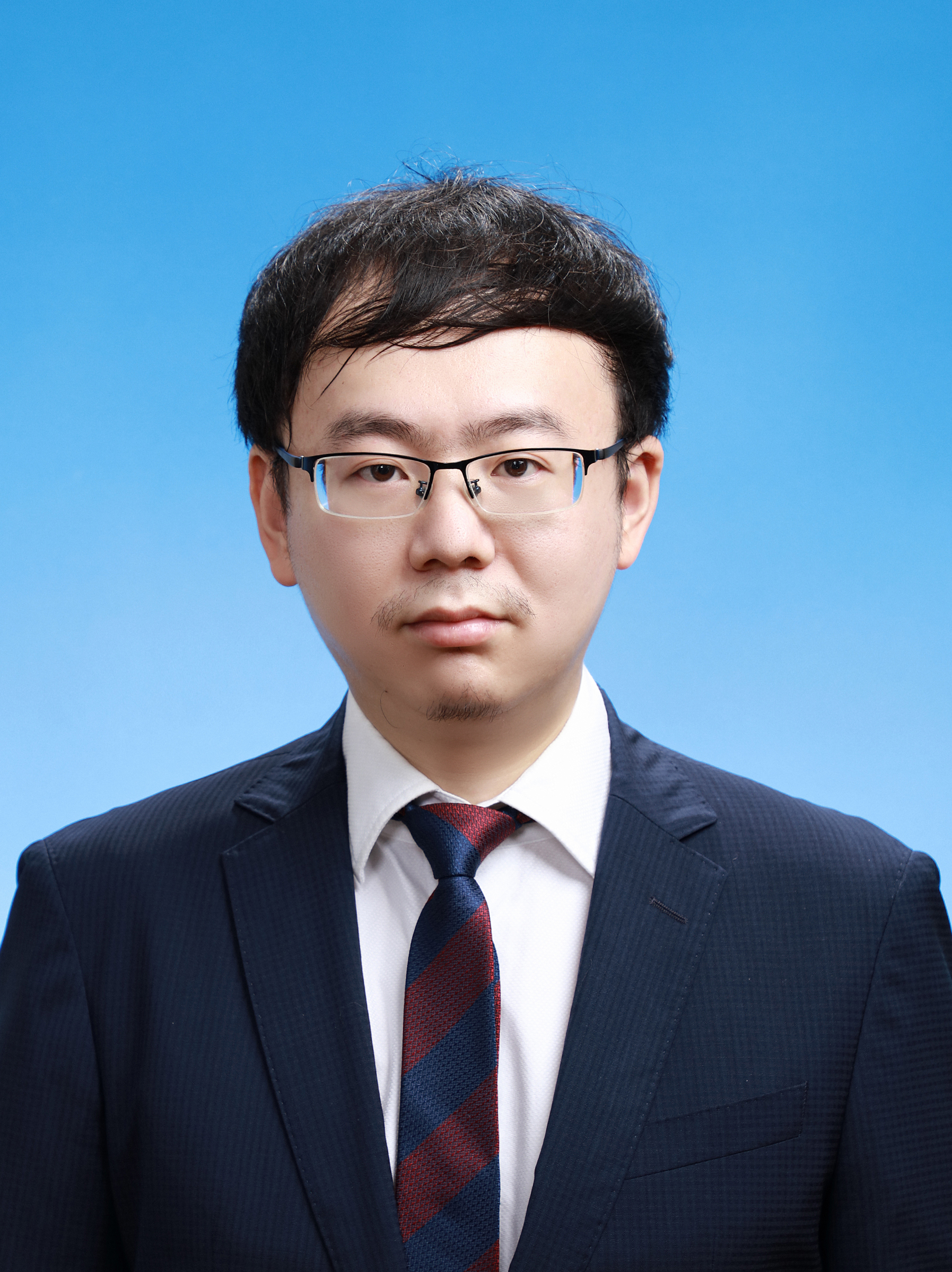}}]{Shaoming He}
	 received the B.Sc. degree and the M.Sc. degree in aerospace engineering from Beijing Institute of Technology, Beijing, China, in 2013 and 2016, respectively, and the Ph.D. degree in aerospace engineering from Cranfield University, Cranfield, U.K., in 2019. He is currently an Associate Professor with School of Aerospace Engineering, Beijing Institute of Technology and also a recognized teaching staff with School of Aerospace, Transport and Manufacturing, Cranfield University. His research interests include aerospace guidance, multitarget tracking and  trajectory optimization. 

	Dr. He received the Lord Kings Norton Medal award from Cranfield University as the most outstanding doctoral student in 2020.
\end{IEEEbiography}

\begin{IEEEbiography}
	[{\includegraphics[width=1in,height=1.25in,clip,keepaspectratio]{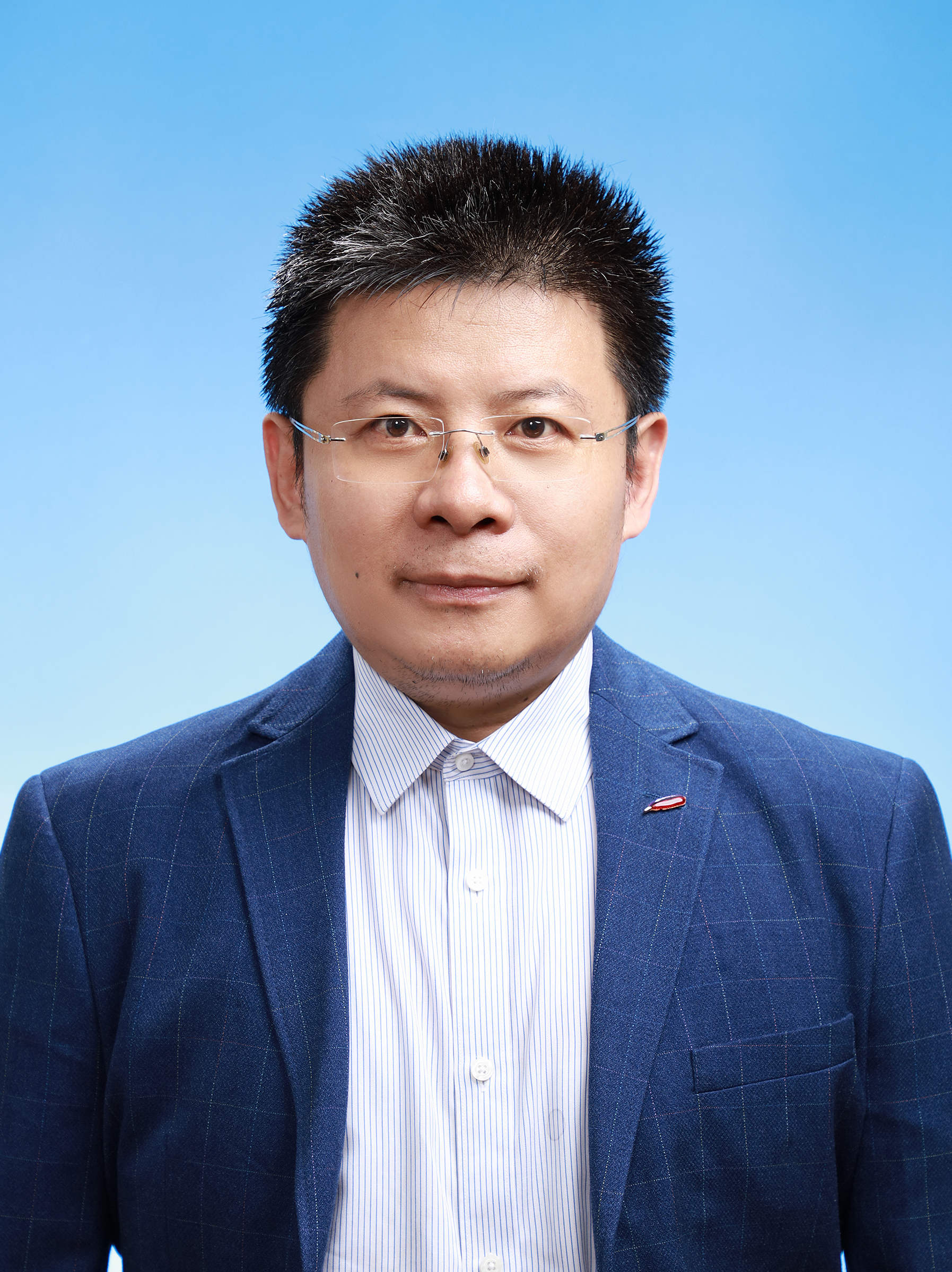}}]{Defu Lin}
	received the M.E. and Ph.D. degrees in Aircraft Design from Beijing Institute of Technology, Beijing, China, in 1999 and 2005, respectively. He is currently a Professor with the School of Aerospace Engineering, Beijing Institute of Technology. He serves as a member of the Unmanned Aircraft Systems Subcommittee of the National Technical Committee for Aircraft Standardization of China. He is the Director of the Beijing Key Laboratory for UAV Autonomous Control and the China-UAE Belt and Road Joint Laboratory on Intelligent Unmanned Systems of Aerospace Engineering, Beijing Institute of Technology. He was selected for the National Talents Program (Ten Thousand Talents Plan) in 2020. His research interests include aircraft system design, flight vehicle guidance, and control technologies.
\end{IEEEbiography}

\newpage
\begin{IEEEbiography}
	[{\includegraphics[width=1in,height=1.25in,clip,keepaspectratio]{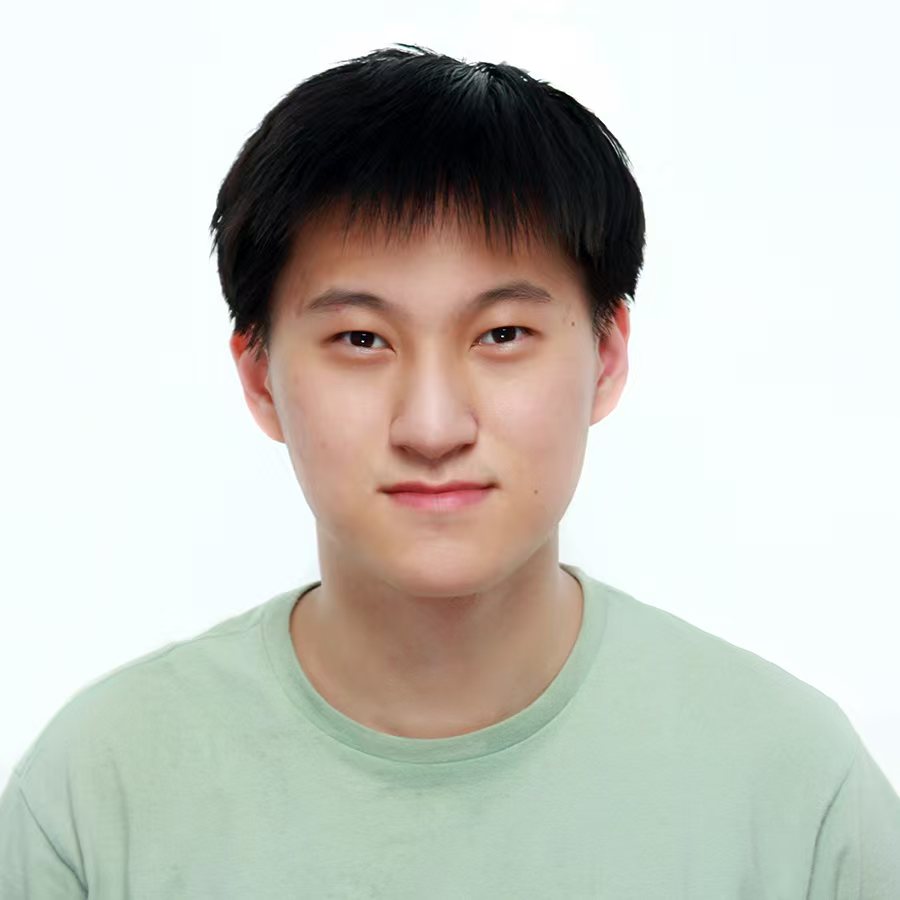}}]{Siqing Cheng}
	Siqing Cheng is currently pursuing the B.Sc. degree in Information and Computing Science at Xi'an Jiaotong-Liverpool University, Suzhou, China. His research interests encompass the application of deep learning to autonomous aerial systems, including real-time small object detection, swarm UAV tracking, and sensor fusion for navigation in GNSS-denied environments.
\end{IEEEbiography}

%\hfill
%\balance

\end{document}